\documentclass[papersize,journal]{IEEEtran}

\usepackage{amsmath,amsfonts}
\usepackage{amssymb}
\usepackage{bbm}
\usepackage{algorithmic}
\usepackage{algorithm}
\usepackage{array}
\usepackage[caption=false,font=normalsize,labelfont=sf,textfont=sf]{subfig}
\usepackage{textcomp}
\usepackage{stfloats}
\usepackage{url}
\usepackage{verbatim}
\usepackage{graphicx}
\usepackage{cite}
\usepackage{booktabs}
\usepackage{multirow}
\usepackage{pifont}
\usepackage{colortbl}
\usepackage{xcolor}
\usepackage{tikz}
\usetikzlibrary{arrows.meta,positioning,fit,calc}
\usepackage{pgfplots}
\pgfplotsset{compat=1.18}
\usepackage[hidelinks]{hyperref}
\newcommand{\etal}{\textit{et al.}}

\begin{document}

\title{Graph-based Semantic Calibration Network for Unaligned UAV RGBT Image Semantic Segmentation and A Large-scale Benchmark}
\author{Fangqiang Fan, Zhicheng Zhao*, Xiaoliang Ma, Chenglong Li, and Jin Tang
\thanks{* Corresponding author: Zhicheng Zhao.}
\thanks{This work was supported in part by the National Natural Science Foundation of China (No. 62306005, 62006002, 62076003, 62376005 and 62576006), and in part by the Natural Science Foundation of Anhui Higher Education Institution (No. 2022AH040014).}
\thanks{F. Fan, Z. Zhao, and C. Li are with Key Laboratory of Intelligent Computing \& Signal Processing (Anhui University), Ministry of Education, Anhui Provincial Key Laboratory of Multimodal Cognitive Computation, School of Artificial Intelligence, Anhui University, Hefei 230601, China.  (Email: fanadmin@163.com, zhaozhicheng@ahu.edu.cn, lcl1314@foxmail.com).}
\thanks{X. Ma is with the School of Computer Science and Technology, Anhui University, Hefei 230601, China, and also with GEOVIS Earth Technology Co., Ltd., Hefei 230088, China.}
\thanks{J. Tang is with the Anhui Provincial Key Laboratory of Multimodal Cognitive Computation, School of Computer Science and Technology, Anhui University, Hefei 230601, China. (Email: tangjin@ahu.edu.cn).}
}

\markboth{Fan \etal: Unaligned UAV RGBT Image Semantic Segmentation}
{Fan \etal: Unaligned UAV RGBT Image Semantic Segmentation}

\maketitle

\begin{abstract}
Fine-grained RGBT image semantic segmentation is crucial for all-weather unmanned aerial vehicle (UAV) scene understanding. 
However, UAV RGBT image semantic segmentation faces two coupled challenges: cross-modal spatial misalignment caused by sensor parallax and platform vibration,
and severe semantic confusion among fine-grained ground objects under top-down aerial views. 
To address these issues, we propose a Graph-based Semantic Calibration Network (GSCNet) for unaligned UAV RGBT image semantic segmentation. 
Specifically, we design a Feature Decoupling and Alignment Module (FDAM) that decouples each modality into shared structural and private perceptual components and performs deformable alignment in the shared subspace, enabling robust spatial correction with reduced modality appearance interference. 
Moreover, we propose a Semantic Graph Calibration Module (SGCM) that explicitly encodes the hierarchical taxonomy and co-occurrence regularities among ground-object categories in UAV scenes into a structured category graph, and incorporates these priors into graph-attention reasoning to calibrate predictions of visually similar and rare categories.
In addition, we construct the Unaligned RGB-Thermal Fine-grained (URTF) benchmark, to the best of our knowledge, the largest and most fine-grained benchmark for unaligned UAV RGBT image semantic segmentation, containing over 25,000 image pairs across 61 semantic categories with realistic cross-modal misalignment. 
Extensive experiments on URTF demonstrate that GSCNet significantly outperforms state-of-the-art methods, with notable gains on fine-grained categories. 
The dataset is available at \url{https://github.com/mmic-lcl/Datasets-and-benchmark-code}.
\end{abstract}


\section{Introduction}

\IEEEPARstart{U}{nmanned} aerial vehicles (UAVs) are widely used for all-weather scene understanding in applications such as urban planning, precision agriculture, and traffic monitoring~\cite{wang2021loveda,lyu2020uavid}. 
Semantic segmentation is a core capability in these applications. RGB images provide rich texture cues but degrade under low illumination and adverse weather, whereas thermal infrared images remain informative under these conditions but provide coarser structural details due to their lower spatial resolution~\cite{sakaridis2018semantic,vertens2020heatnet}. Given these complementary strengths, fusing the two modalities is a natural choice for robust UAV scene understanding~\cite{ha2017mfnet,shivakumar2020pst900}. 
However, on real dual-sensor UAV platforms, RGB and thermal images are rarely pixel-aligned because sensor parallax and platform vibration introduce spatially varying offsets. 
Unaligned UAV RGBT image semantic segmentation aims to predict semantic masks from spatially misaligned RGB-Thermal image pairs under realistic UAV sensing conditions.

\begin{figure}[!t]
    \centering
    \includegraphics[width=\linewidth]{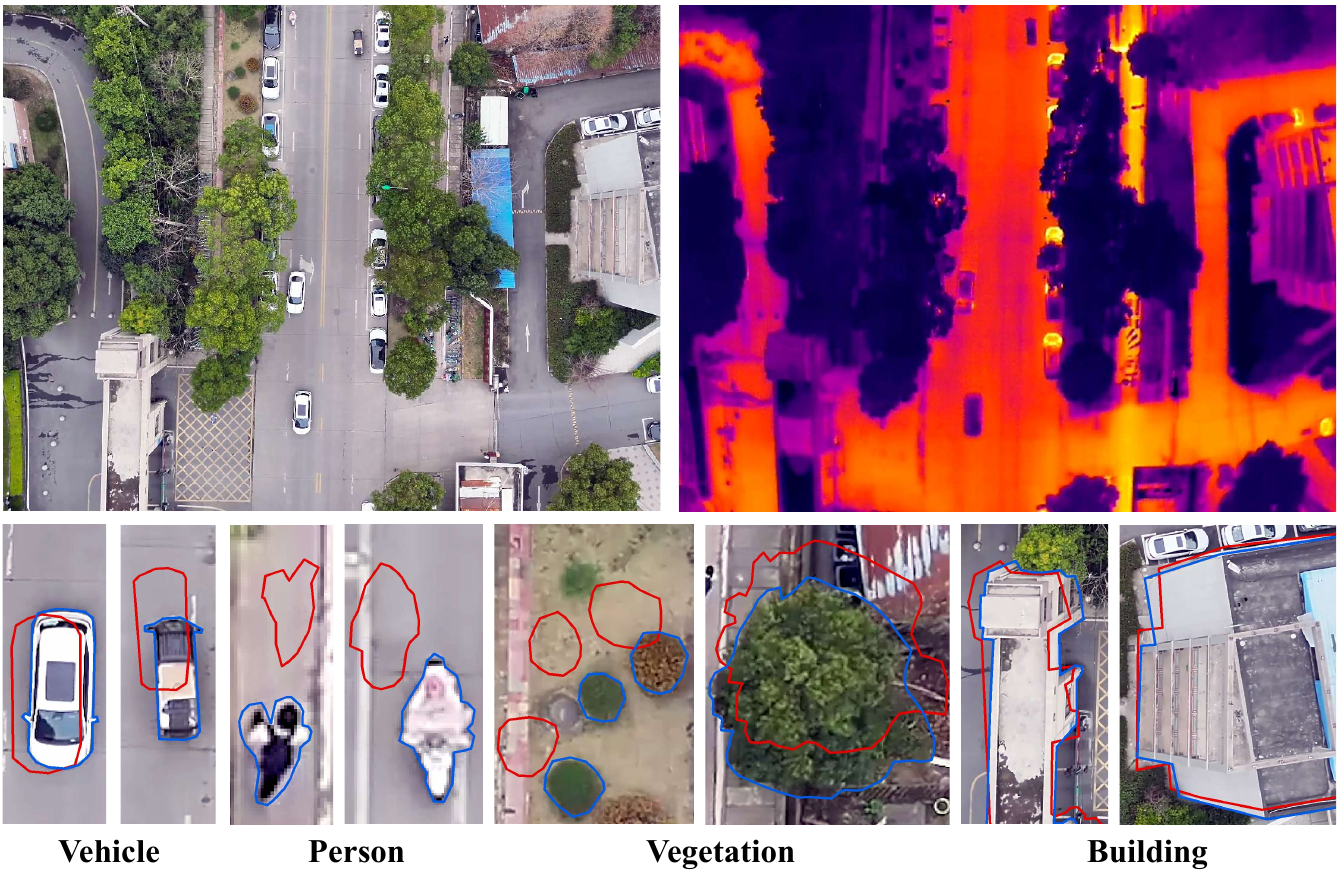}
    \caption{Cross-modal boundary misalignment in RGB-Thermal UAV imaging. The top row shows an RGB image and its thermal counterpart. In the bottom row, blue contours denote object boundaries in the RGB modality and red contours denote the corresponding boundaries in the thermal modality. Representative examples from vehicles, persons, vegetation, and buildings show that cross-modal spatial offsets are widespread in UAV scenes.}
    \label{fig:motivation1}
\end{figure}

Cross-modal spatial misalignment is difficult to handle in this setting because the offsets are spatially varying and often object-dependent, so no single global transformation can remove them.
As Fig.~\ref{fig:motivation1} shows, such boundary discrepancies occur across diverse object categories in UAV scenes.
When methods such as CMX~\cite{zhang2023cmx} fuse these misaligned features under an implicit alignment assumption, they mix responses from semantically inconsistent locations, producing ghosting artifacts, blurred boundaries, and missed small targets~\cite{chen2024weakly}.

\begin{figure}[t]
    \centering
    \includegraphics[width=\linewidth]{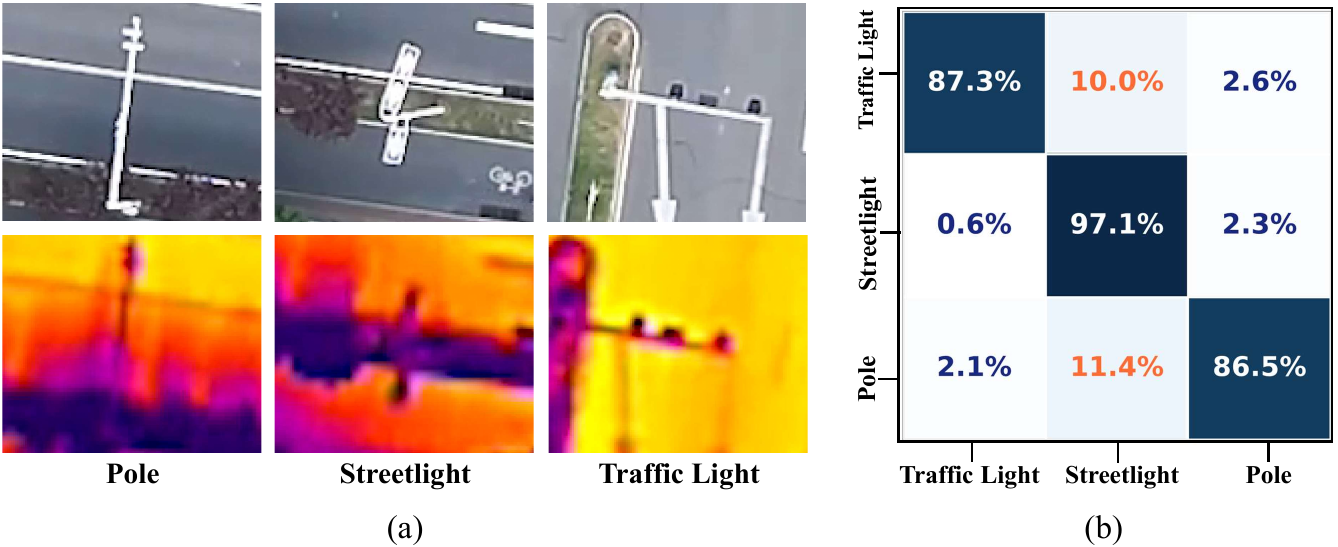}
    \caption{Fine-grained semantic confusion in UAV aerial scenes. (a) Pole, streetlight, and traffic light occupy only a small number of pixels in both RGB and thermal images and exhibit highly similar appearance from the UAV viewpoint, making them difficult to distinguish. (b) Confusion matrix of these three categories produced by AMDANet, where traffic light and pole are frequently misclassified as streetlight.}
    \label{fig:motivation2}
\end{figure}

Beyond spatial misalignment, severe semantic confusion among fine-grained ground objects poses another major challenge. Under top-down aerial views, many categories share similar visual appearance. Poles, streetlights, and traffic lights, for instance, occupy few pixels and look alike in both RGB and thermal images. The pixel distribution is also long-tailed, leaving rare classes with too few samples to learn stable boundaries. Most existing methods classify pixels in a flat label space and ignore hierarchical or co-occurrence regularities among related categories. As Fig.~\ref{fig:motivation2} shows, these factors make categories such as pole, streetlight, and traffic light hard to distinguish, with tail classes suffering most from limited training pixels.

These two problems interact: misalignment corrupts the visual evidence that fine-grained recognition depends on, which in turn amplifies confusion among categories that already look alike.
We propose GSCNet, a Graph-based Semantic Calibration Network that integrates spatial alignment and semantic calibration into a unified end-to-end framework. Estimating offsets directly in the raw feature space is unreliable because RGB and thermal features differ in appearance and contrast. Our Feature Decoupling and Alignment Module (FDAM) therefore first decouples each modality into shared structural and private perceptual components and then estimates deformable offsets in the shared subspace where the modality gap is reduced, deriving geometric corrections from structurally consistent representations without discarding modality-specific cues.
Fine-grained semantic confusion is hard to resolve through a flat label space and local visual evidence alone, because visually similar categories lack discriminative cues at the pixel level and rare classes have too few samples to learn reliable boundaries. Our Semantic Graph Calibration Module (SGCM) encodes hierarchical taxonomy and co-occurrence regularities among ground-object categories into a structured category graph and uses graph-attention reasoning so that visually similar and rare categories can borrow discriminative context from semantically related nodes.

In addition, to promote research on unaligned UAV RGBT image semantic segmentation, we construct a large-scale benchmark named Unaligned RGB-Thermal Fine-grained (URTF).
Ground-level RGB-Thermal datasets such as MFNet~\cite{ha2017mfnet}, PST900~\cite{shivakumar2020pst900}, FMB~\cite{liu2023multi}, and MVSeg~\cite{zhang2024mrfs} assume strict pixel-wise registration and provide only coarse category definitions, while UAV-specific datasets such as CART~\cite{chen2024cart}, MVUAV~\cite{ji2024unleashing}, and Kust4K~\cite{ouyang2025kust4k} still offer limited category granularity or rely on pixel-level alignment.
With over 25,000 image pairs spanning 61 semantic categories under realistic cross-modal misalignment and diverse illumination and weather conditions, URTF is the largest and most fine-grained benchmark currently available for this setting.

The primary contributions of this work are summarized as follows:
\begin{itemize}
\item We construct URTF, to the best of our knowledge the largest and most fine-grained benchmark for unaligned UAV RGBT image semantic segmentation, containing over 25,000 image pairs across 61 semantic categories with realistic cross-modal misalignment under diverse illumination and weather conditions.
\item We propose GSCNet, a unified spatial-semantic framework for robust fine-grained unaligned UAV RGBT image semantic segmentation, which jointly addresses cross-modal spatial misalignment and fine-grained semantic confusion.
\item We propose the Feature Decoupling and Alignment Module~(FDAM), which decouples RGB-Thermal features into shared structural and private perceptual components for illumination-aware deformable alignment. In addition, we introduce the Semantic Graph Calibration Module~(SGCM), which explicitly encodes hierarchical taxonomy and co-occurrence regularities into a structured category graph and calibrates predictions via graph-attention reasoning.
\item Extensive quantitative and qualitative experiments on URTF demonstrate that GSCNet significantly outperforms existing state-of-the-art RGBT image semantic segmentation methods, with notable gains on fine-grained categories under challenging UAV sensing conditions.
\end{itemize}

\section{Related Work}

\subsection{RGBT Semantic Segmentation}

Early multimodal segmentation methods adopt dual-stream CNNs with element-wise fusion. FuseNet~\cite{hazirbas2016fusenet}, originally designed for RGB-D, establishes the element-wise summation paradigm later widely adopted in RGBT segmentation. MFNet~\cite{ha2017mfnet} introduces a lightweight mini-inception encoder for real-time RGBT parsing, and RTFNet~\cite{sun2019rtfnet} progressively folds thermal features into the RGB decoder. These encoder-fusion designs establish the basic dual-stream paradigm but rely on hand-crafted aggregation rules that cannot selectively weight informative regions.
Subsequent work introduces attention mechanisms to modulate cross-modal contributions. FEANet~\cite{deng2021feanet} applies a feature-enhanced attention module to exploit fine spatial details, EGFNet~\cite{zhou2022edge} adds edge-guided attention for boundary refinement, and GMNet~\cite{zhou2022gmnet} exploits graded multimodal features through a multilabel-learning framework for urban scene parsing. More recently, Transformer-based architectures further extend the fusion receptive field: CMX~\cite{zhang2023cmx} designs cross-modal feature rectification and feature fusion modules within a unified RGB-X framework, achieving state-of-the-art results on multiple benchmarks.
Another line of work jointly optimizes image fusion and semantic segmentation. MRFS~\cite{zhang2024mrfs} formulates the two tasks in a shared multi-interactive feature learning framework, and AMDANet~\cite{Zhong_2025_ICCV} mitigates cross-modal feature discrepancies through attention-driven multi-perspective alignment for joint RGB-infrared fusion and segmentation. Meanwhile, architectures originally designed for other multimodal settings have also been adapted to RGBT segmentation. DFormerV2~\cite{yin2025dformerv2} injects depth-derived geometry priors into self-attention for RGB-D segmentation, and MambaSeg~\cite{wang2026mambaseg} explores Mamba-based state-space modeling for efficient RGB-event dense prediction. These methods advance fusion quality or architectural generality, yet they still assume well-registered input pairs.
Alongside architectural progress, benchmarks have expanded from the 9-class MFNet dataset~\cite{ha2017mfnet} to 36-category UAV benchmarks~\cite{ji2024unleashing}, and UAV-specific datasets such as CART~\cite{chen2024cart} and Kust4K~\cite{ouyang2025kust4k} have also emerged. U-MFNet~\cite{zhou2025drgbt} constructs synthetically deformed pairs from MFNet to study unregistered fusion, but the deformations are artificial and the label space remains limited to 9 classes. Despite this progress, no RGB-Thermal benchmark simultaneously provides fine-grained annotation granularity, large-scale coverage, and realistic cross-modal misalignment.

\subsection{Fine-Grained Semantic Segmentation}

Semantic segmentation has advanced from FCN-based encoders~\cite{long2015fully,chen2017deeplab} toward fine-grained recognition, yet global context mechanisms such as dilated convolutions~\cite{yu2015multi} and ASPP~\cite{chen2018deeplabv3plus} enlarge receptive fields without modeling explicit inter-class relationships, making them insufficient to disambiguate subcategories that share similar local appearance but differ in semantic identity. Fine-grained segmentation faces two coupled difficulties: inter-class confusion among visually similar subcategories and long-tailed recognition where rare categories are under-optimized.
Graph-based reasoning has been introduced to capture inter-class relations beyond local receptive fields. GloRe~\cite{chen2019glore} projects pixel features onto a set of latent nodes and performs relational reasoning in the graph domain, while DGMN~\cite{zhang2020dgmn} dynamically generates graph structures conditioned on each input image. In remote sensing, SAGRNet~\cite{gui2025sagrnet} introduces an object-based graph convolutional network with sampling aggregation and self-attention for vegetation cover classification. Other methods leverage external semantic priors: hierarchy-aware losses~\cite{bertinetto2020hierarchy} penalize predictions according to hierarchical class distance, and label co-occurrence graphs~\cite{chen2019mlgcn} model inter-label dependencies for multi-label recognition. However, purely data-driven graphs lack interpretable structure, whereas static prior graphs cannot adapt to scene-specific category distributions.
Our SGCM combines both: it initializes the adjacency from hierarchical and co-occurrence priors and augments it with a learnable residual that adapts to data-driven patterns during training.

\subsection{Unaligned Multimodal Fusion}

Estimating spatial correspondence between heterogeneous sensor modalities is a prerequisite for coherent feature fusion. Classical global transforms~\cite{detone2016deep} cannot capture local, depth-dependent parallax on UAV platforms, while parametric spatial transformers~\cite{jaderberg2015spatial} and dense optical-flow methods~\cite{dosovitskiy2015flownet,sun2018pwc} can model geometric transformations but suffer from the modality-gap dilemma in which appearance discrepancy is conflated with genuine spatial offsets; deformable convolution~\cite{zhu2019deformable} adapts sampling locations for intra-modal irregularities but does not address unreliable cross-modal offset estimation.
Recent work couples alignment with downstream tasks. OAFA~\cite{chen2024weakly} projects RGB-Thermal features into a common subspace for deformable offset estimation, and RegSeg~\cite{tip_e2e_reg_seg} jointly optimizes registration and segmentation through a shared encoder. These methods improve alignment but do not explicitly disentangle modality-shared structural cues from modality-private perceptual cues, leaving alignment exposed to residual cross-modal interference. Shared-private decomposition methods~\cite{hazarika2020misa,xu2020learning} reduce inter-modal discrepancy but do not recover spatial correspondence.
Our FDAM bridges these two lines: it performs explicit shared-private decomposition with contrastive and orthogonality constraints, estimates deformable offsets in the modality-shared structural subspace, and introduces illumination-adaptive anchor selection to handle day-night variation.

\section{Method}

\subsection{Overall Architecture}

\begin{figure*}[!t]
    \centering
    \includegraphics[width=\linewidth]{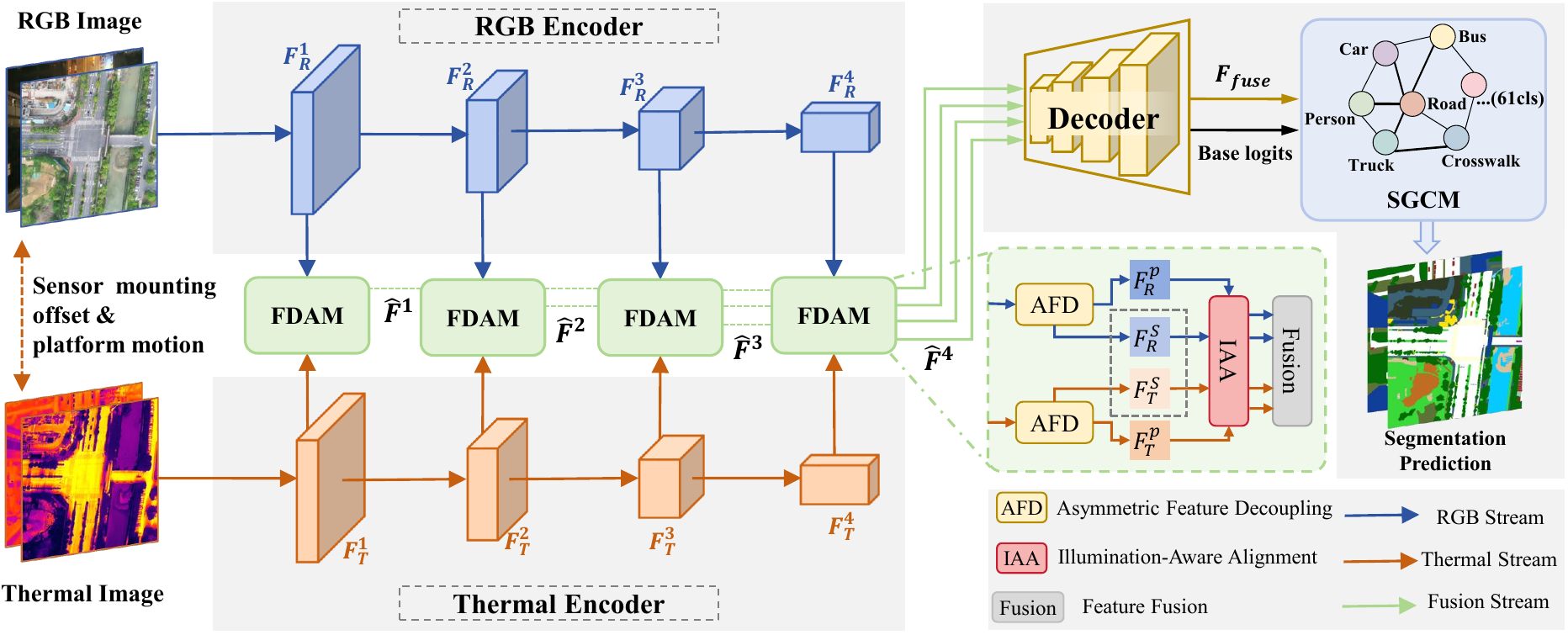}
    \caption{Overview of GSCNet. RGB and thermal images are processed by two modality-specific MiT-B4 encoders, and FDAM is inserted at all four stages for cross-modal feature interaction. Within FDAM, AFD decomposes features into shared structural and private perceptual components, while IAA performs illumination-aware bidirectional deformable alignment in the shared subspace. The aligned multi-scale features are fused by the decoder to produce $\mathbf{F}_{\mathrm{fuse}}$ and initial logits $\mathbf{L}_0$, which are further calibrated by SGCM via graph-attention reasoning over a structured category graph with hierarchical and co-occurrence regularities to obtain the final segmentation.}
    \label{fig:framework}
\end{figure*}

Fig.~\ref{fig:framework} shows the overall architecture. GSCNet builds on SegFormer~\cite{xie2021segformer} with two modality-specific Mix Transformer (MiT) branches that generate four-stage feature hierarchies ($C_i \in \{64, 128, 320, 512\}$). The two branches share the same architecture but use independent parameters to accommodate appearance differences between modalities. At each stage, FDAM decouples the features into shared structural and private perceptual components and aligns them in the shared subspace. The SegFormer all-MLP decoder then aggregates the aligned multi-scale features to produce the fused representation $\mathbf{F}_{\mathrm{fuse}}$ and the base logits $\mathbf{L}_0$.
FDAM handles cross-modal spatial misalignment at the feature level; SGCM then calibrates $\mathbf{L}_0$ through graph-attention reasoning over a structured category graph encoding hierarchical and co-occurrence regularities. Sections~\ref{sec:fdam} and~\ref{sec:sgcm} describe each module in detail.

\subsection{Feature Decoupling and Alignment Module (FDAM)}
\label{sec:fdam}

Applying deformable convolutions~\cite{zhu2019deformable} directly to raw multimodal features is unreliable: the offset predictor confuses genuine spatial displacements with the inherent appearance gap between RGB texture and thermal radiation. FDAM addresses this with a decouple-then-align strategy inspired by shared-private representation learning~\cite{xu2020learning} and cross-modal alignment studies~\cite{wang2019rgb}. Instead of assuming that raw RGB and thermal features are directly comparable, it first separates each modality into a shared structural branch and a private perceptual branch via AFD, estimates deformable offsets in the shared subspace via IAA, and reuses the same geometric corrections for the private branches under illumination-adaptive anchor selection.

\subsubsection{Asymmetric Feature Decoupling (AFD)}

As shown in Fig.~\ref{fig:fdam1}(a), AFD decomposes each stage into one shared encoder and two private encoders. The shared encoder $\phi^s$ is a lightweight two-layer Conv-BN-ReLU block whose weights are shared across the RGB and thermal streams, encouraging both modalities to meet in a common structural subspace. In contrast, each private encoder is a shallower modality-specific single-layer block, so it mainly retains sensory details such as RGB texture and thermal intensity patterns.
Formally, for stage $i$, the decoupling process is defined as:
\begin{equation}
\begin{aligned}
  F_R^{s(i)} &= \phi^s(F_R^{(i)}), \qquad F_T^{s(i)} = \phi^s(F_T^{(i)}), \\
  F_R^{p(i)} &= \phi_R^p(F_R^{(i)}), \quad\; F_T^{p(i)} = \phi_T^p(F_T^{(i)}),
\end{aligned}
\end{equation}
where $\phi^s$ denotes the modality shared encoder, and $\phi_R^p$, $\phi_T^p$ denote the modality private encoders.

Inspired by shared-private disentanglement in multimodal representation learning~\cite{hazarika2020misa}, we train AFD with three task-specific constraints. The primary objective is a patch-based contrastive alignment loss $\mathcal{L}_{\mathrm{align}}^{(i)}$ that pulls corresponding structural patches together while separating mismatched ones:
\begin{equation}
  \mathcal{L}_{\mathrm{align}}^{(i)} = -\frac{1}{BN_i} \sum_{b,j}
  \log \frac{\exp\!\bigl(\hat{P}_{R,b,j}^{\top} \hat{P}_{T,b,j}\,/\,\tau\bigr)}
  {\sum_{k} \exp\!\bigl(\hat{P}_{R,b,j}^{\top} \hat{P}_{T,b,k}\,/\,\tau\bigr)},
\end{equation}
where $\hat{P}_R, \hat{P}_T \in \mathbb{R}^{B \times N_i \times C_i}$ are $\ell_2$-normalized patch vectors obtained by average pooling with kernel size 8, yielding $N_i = \lfloor H_i/8 \rfloor \times \lfloor W_i/8 \rfloor$ patches per stage, and $\tau = 0.07$. The combined downsampling of the encoder ($2^{i+1}$) and the pooling stride subsumes residual cross-modal offsets within each patch, so $\mathcal{L}_{\mathrm{align}}$ enforces structural correspondence rather than pixel-level alignment.
Without further regularization, the shared and private branches may converge to redundant representations. We therefore add an orthogonality loss $\mathcal{L}_{\mathrm{orth}}^{(i)}$, defined as the squared mean per-pixel cosine similarity between $F_m^{s(i)}$ and $F_m^{p(i)}$ for each modality $m$, to keep the two subspaces well separated, and an auxiliary segmentation loss $\mathcal{L}_{\mathrm{sem}}^{(i)}$, computed from a lightweight head supervised by ground-truth labels downsampled to each stage's resolution, to anchor the shared features to task-relevant structure.

\subsubsection{Illumination-Aware Alignment (IAA)}

\begin{figure}[!t]
  \centering
  \includegraphics[width=\linewidth]{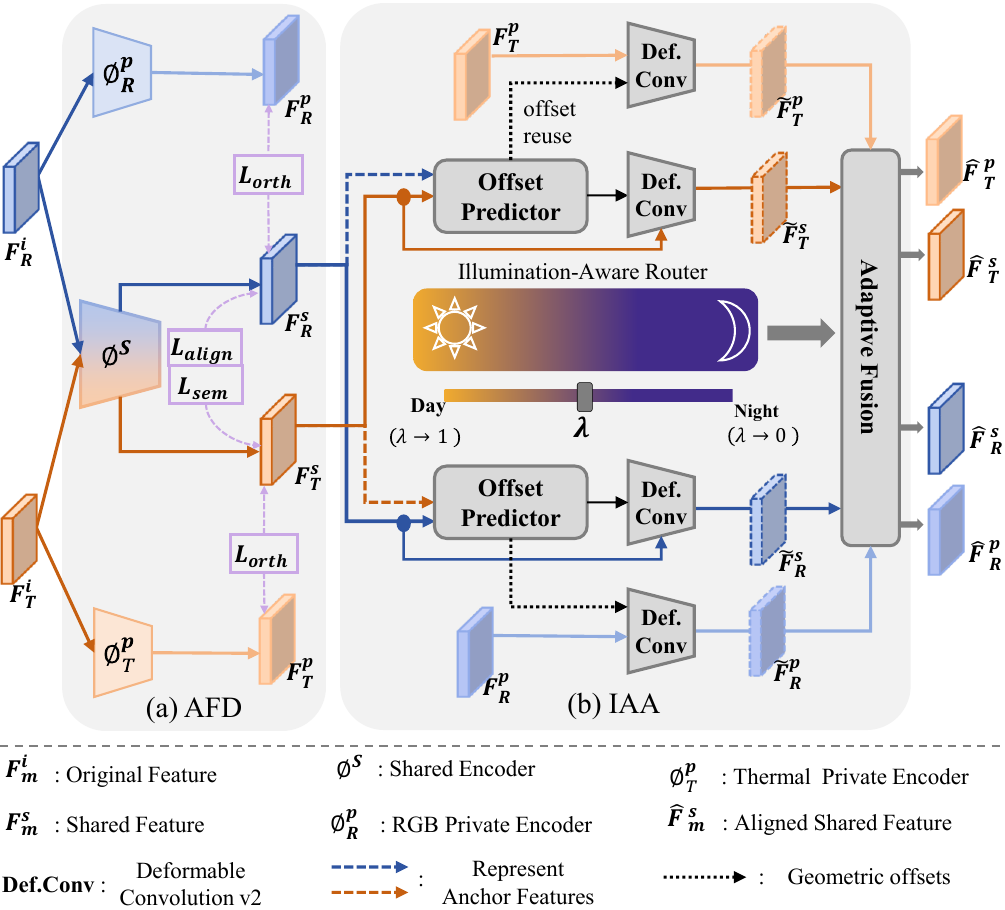}
  \caption{Overview of FDAM. (a) Asymmetric Feature Decoupling (AFD): separates each modality's features into shared structural and private perceptual components. (b) Illumination-Aware Alignment (IAA): bidirectional deformable alignment with adaptive anchor selection guided by illumination weight $\lambda$. }
  \label{fig:fdam1}
\end{figure}

AFD reduces the modality gap but leaves residual geometric offsets. As illustrated in Fig.~\ref{fig:fdam1}(b), IAA corrects these offsets through bidirectional offset estimation, deformable warping, and illumination-aware anchor selection.
At each stage $i$, two lightweight offset predictors first estimate the warps for the two possible alignment directions: $\mathcal{D}_f^{(i)}$ keeps RGB fixed and warps thermal toward it, whereas $\mathcal{D}_b^{(i)}$ keeps thermal fixed and warps RGB toward it:
\begin{equation}
\begin{aligned}
  (\Delta \mathbf{p}_f^{(i)}, \mathbf{m}_f^{(i)}) &= \mathcal{D}_f^{(i)}\!\left(F_R^{s(i)} \,\|\, F_T^{s(i)}\right),\\
  (\Delta \mathbf{p}_b^{(i)}, \mathbf{m}_b^{(i)}) &= \mathcal{D}_b^{(i)}\!\left(F_T^{s(i)} \,\|\, F_R^{s(i)}\right).
\end{aligned}
\end{equation}
Each predictor specializes in a single warp direction, reducing the complexity of offset estimation.

We apply the estimated offsets via deformable convolutions v2 (DCNv2)~\cite{zhu2019deformable} to warp the shared structural features:
\begin{equation}
\begin{aligned}
  \tilde{F}_T^{s(i)} &= \mathrm{DCNv2}\!\left(F_T^{s(i)},\;\Delta \mathbf{p}_f^{(i)},\;\mathbf{m}_f^{(i)}\right),\\
  \tilde{F}_R^{s(i)} &= \mathrm{DCNv2}\!\left(F_R^{s(i)},\;\Delta \mathbf{p}_b^{(i)},\;\mathbf{m}_b^{(i)}\right).
\end{aligned}
\end{equation}
A fixed reference modality fails across illumination changes: RGB provides sharper boundaries in daytime, while thermal provides more reliable structure at night. IAA uses a lightweight router to predict a global image-level illumination-aware weight $\lambda \in [0,1]$, shared by all stages, through $\lambda = \sigma(\mathrm{MLP}(\mathrm{GAP}(\mathbf{I}_{\mathrm{rgb}})))$, where $\mathrm{GAP}$ is global average pooling and the two-layer $\mathrm{MLP}$ has hidden dimension 16 with fewer than 1K parameters. The warped and original features are softly blended according to $\lambda$:
\begin{equation}
\begin{aligned}
  \hat{F}_T^{s(i)} &= \lambda\cdot\tilde{F}_T^{s(i)} + (1-\lambda)\cdot F_T^{s(i)},\\
  \hat{F}_R^{s(i)} &= (1-\lambda)\cdot\tilde{F}_R^{s(i)} + \lambda\cdot F_R^{s(i)}.
\end{aligned}
\end{equation}
When $\lambda\!\to\!1$, thermal features are aligned to the RGB reference frame; when $\lambda\!\to\!0$, RGB features are aligned to the thermal reference frame. The fused representation thus adopts whichever modality is structurally more trustworthy under the current illumination as the spatial anchor.
Because the decoder later combines shared and private features, the private branches should reside in the same reference frame. We therefore reuse the same geometric offsets $(\Delta \mathbf{p}_f^{(i)}, \mathbf{m}_f^{(i)})$ and $(\Delta \mathbf{p}_b^{(i)}, \mathbf{m}_b^{(i)})$ to warp $F_T^{p(i)}$ and $F_R^{p(i)}$ via DCNv2 and apply the same $\lambda$-blending to obtain the aligned private features $\hat{F}_T^{p(i)}$ and $\hat{F}_R^{p(i)}$, while keeping modality-specific learnable DCN weights $\mathbf{W}_f^{p(i)}$ and $\mathbf{W}_b^{p(i)}$ (initialized as center-one identity kernels) to account for their different appearance statistics.

\subsubsection{Feature Fusion}

Inspired by disentangled cross-modal fusion~\cite{chen2020disentangled}, we concatenate and compress the aligned shared features from both modalities, then append the two private branches to form the stage-wise fused feature:
\begin{equation}
  \hat{F}^{(i)} = \mathrm{Conv}_{1\times1}\!\left(\hat{F}_R^{s(i)} \,\|\, \hat{F}_T^{s(i)}\right) \,\|\, \hat{F}_R^{p(i)} \,\|\, \hat{F}_T^{p(i)},
\end{equation}
where $\|$ denotes channel-wise concatenation and $\mathrm{Conv}_{1\times1}$ halves the channel dimension to keep the representation compact. The decoder aggregates $\{\hat{F}^{(i)}\}_{i=1}^{4}$ across all four stages and produces the fused feature map $\mathbf{F}_{\mathrm{fuse}}$ and base logits $\mathbf{L}_0$.

\subsection{Semantic Graph Calibration Module (SGCM)}
\label{sec:sgcm}

After FDAM reduces spatial offsets, the dominant remaining errors are semantic. Under top-down UAV views, many ground-object categories share similar local appearance and occupy comparable spatial extents, yet standard convolutions that aggregate nearby cues cannot capture their inter-class semantic relations. Tail categories suffer further because they provide too few pixels to learn stable decision boundaries from local context alone. Motivated by graph reasoning networks~\cite{chen2019glore,zhang2020dgmn}, SGCM builds a learnable category graph from $\mathbf{F}_{\mathrm{fuse}}$ and $\mathbf{L}_0$, using hierarchical taxonomy and co-occurrence priors to calibrate the base logits.

\subsubsection{Category Graph and Prior Adjacency}
\label{sec:sgcm-graph}

We define the category graph as $\mathcal{G}=(\mathcal{V},\mathcal{E},\tilde{\mathbf{A}})$, where $\mathcal{V}=\{v_1,\dots,v_K\}$ corresponds to the $K$ semantic categories and $\tilde{\mathbf{A}}\in\mathbb{R}^{K\times K}$ is the prior-guided adjacency defined below. We first describe how the edges are constructed from category-level relational priors that are otherwise hard to learn from local appearance alone; the image-specific node features $\mathbf{H}^{(0)}$ are then defined in Sec.~\ref{sec:sgcm-agg}.

As illustrated in Fig.~\ref{fig:sgcm}(a), we initialize the edges with two complementary priors to provide a structured starting point for learning.
\textit{Hierarchical similarity} ($\mathbf{A}_H$). Inspired by hierarchy-aware recognition~\cite{bertinetto2020hierarchy}, we construct a taxonomy-based similarity prior. We organize the 61 categories into a three-level taxonomy with 6 top-level groups and define $\mathbf{A}_H(i,j) = \exp(-d(i,j)/s)$, where $d(i,j)$ is the tree path distance and $s = 2.0$. Categories sharing a common parent (e.g., traffic sign, streetlight, traffic light) receive strong edge weights, enabling rare categories to propagate context through their taxonomic neighbors during graph reasoning.
\textit{Contextual co-occurrence} ($\mathbf{A}_C$). Following the idea of label-dependency modeling from co-occurrence~\cite{chen2019mlgcn}, we define a dataset-specific co-occurrence prior by encoding how often categories appear together in training images: $\mathbf{A}_C(i,j) = N(C_i \cap C_j) / \max(N(C_i), N(C_j))$, where $N(C_i)$ counts training images containing category $i$. The normalization suppresses head-category dominance while preserving scene-level contextual compatibility.

\begin{figure}[t]
  \centering
  \includegraphics[width=.70\linewidth]{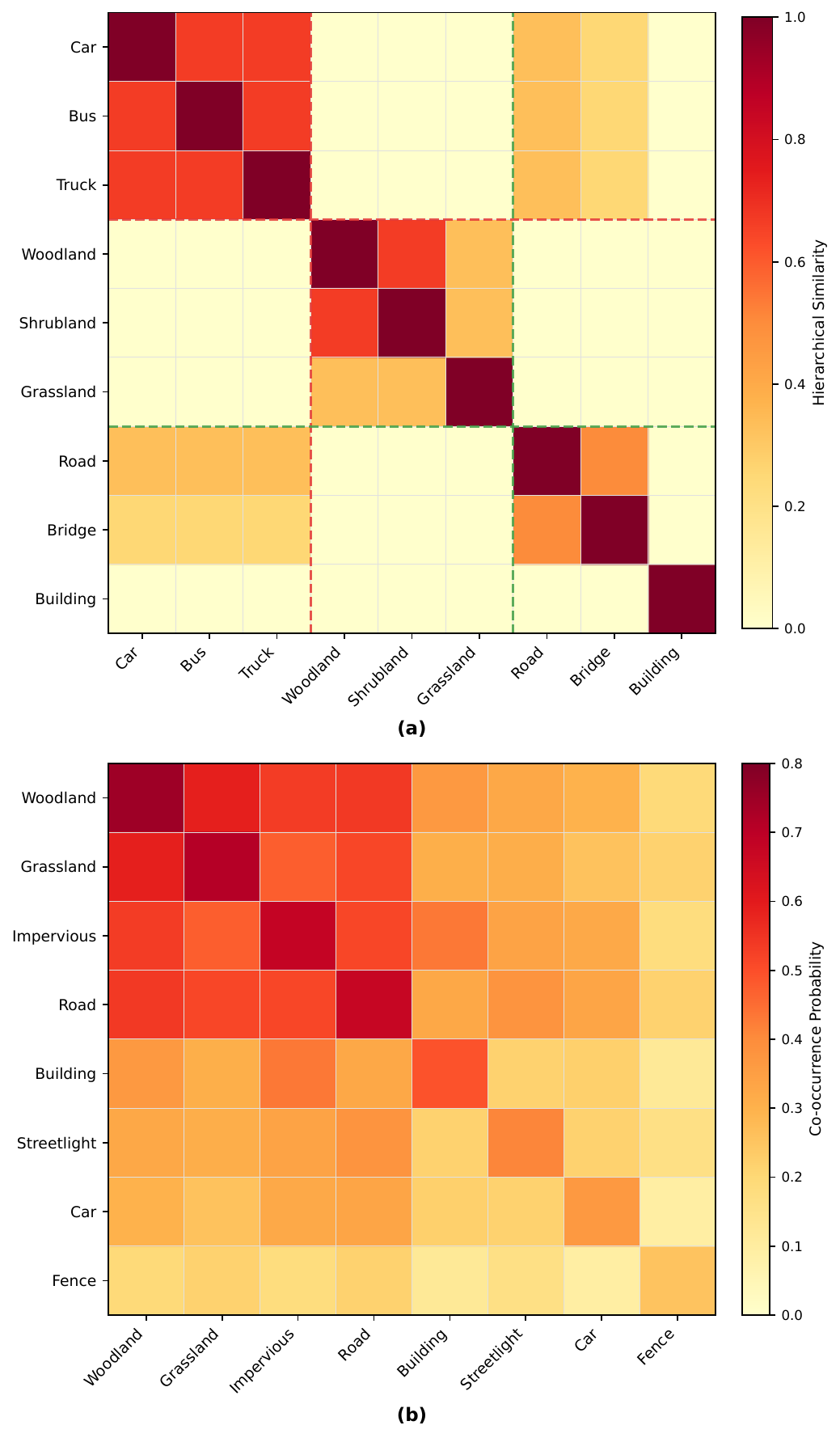}
  \caption{Visualization of the two static prior matrices in SGCM for representative categories. (a)~Hierarchical similarity ($\mathbf{A}_H$): block-diagonal structure reflects taxonomic groupings. (b)~Co-occurrence probability ($\mathbf{A}_C$): encodes scene-level contextual compatibility.}
  \label{fig:prior_heatmaps}
\end{figure}

Fig.~\ref{fig:prior_heatmaps} visualizes the two sources: $\mathbf{A}_H$ exhibits a clustered structure where intra-group weights are large and inter-group weights are near zero, mirroring the taxonomy; $\mathbf{A}_C$ captures scene-level co-occurrence regularities that cut across taxonomic boundaries. We combine them as $\mathbf{A}_{\mathrm{raw}} = 0.6\mathbf{A}_H + 0.4\mathbf{A}_C$ and then symmetrize and normalize the result ($\bar{\mathbf{A}}=0.5(\mathbf{A}_{\mathrm{raw}}+\mathbf{A}_{\mathrm{raw}}^\top)$, $\mathbf{A}_p = \mathbf{D}^{-1/2}\bar{\mathbf{A}}\mathbf{D}^{-1/2}$, where $\mathbf{D}$ is the degree matrix) to obtain the prior adjacency. Static priors cannot cover all inter-class correlations in the training data. For instance, visually similar but taxonomically distant categories may still confuse the classifier. We add a learnable residual $\mathbf{A}_{\delta} \in \mathbb{R}^{K \times K}$, initialized to zero, that strengthens such data-driven relations during training. The final adjacency is:
\begin{figure*}[!t]
  \centering
  \includegraphics[width=\linewidth]{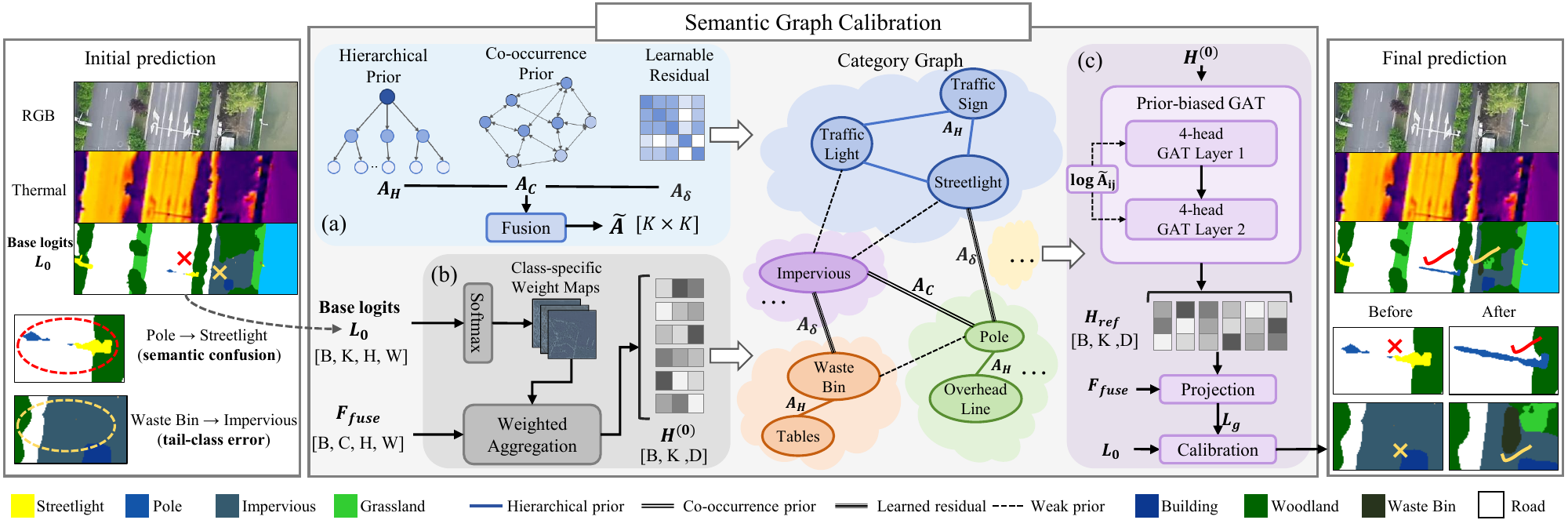}
\caption{Overview of SGCM. Left: the initial prediction exhibits two typical failure modes: semantic confusion (Pole$\to$Streetlight) and tail-class misclassification (Waste Bin$\to$Impervious). (a)~The learnable adjacency $\tilde{\mathbf{A}}$ is built from $\mathbf{A}_H$, $\mathbf{A}_C$, and $\mathbf{A}_{\delta}$; the category graph shows representative edge types: hierarchical (e.g., Traffic Light--Streetlight), co-occurrence (e.g., Pole--Impervious), and learned residual (e.g., Waste Bin--Impervious). (b)~Base logits $\mathbf{L}_0$ generate class-specific weight maps for aggregating node features $\mathbf{H}^{(0)}$ from $\mathbf{F}_{\mathrm{fuse}}$. (c)~A prior-biased GAT refines node embeddings and projects them back to produce graph logits $\mathbf{L}_g$, which are residually fused with $\mathbf{L}_0$ for final prediction. Right: the calibrated prediction corrects both error types.}
  \label{fig:sgcm}
\end{figure*}

\begin{equation}
\label{eq:prior_adj}
  \begin{gathered}
    \mathbf{A}_{\mathrm{upd}} = \mathrm{ReLU}(\mathbf{A}_p + \mathbf{A}_{\delta}), \\
    \tilde{\mathbf{A}} = \mathrm{SymNorm}\!\left(0.5(\mathbf{A}_{\mathrm{upd}} + \mathbf{A}_{\mathrm{upd}}^{\top})\right),
  \end{gathered}
\end{equation}
where $\mathrm{SymNorm}(\cdot) = \mathbf{D}^{-1/2}(\cdot)\mathbf{D}^{-1/2}$ denotes symmetric degree normalization as above.
An $\ell_1$ penalty $\mathcal{L}_{\mathrm{kg}}$ (Section~\ref{sec:loss}) keeps $\mathbf{A}_{\delta}$ sparse, so the model can adapt the prior without drifting too far from the dataset structure.

\subsubsection{Image-Specific Dynamic Aggregation}
\label{sec:sgcm-agg}

As shown in Fig.~\ref{fig:sgcm}(b), SGCM initializes every node from the current image through soft attention pooling. The base logits $\mathbf{L}_0$ are converted into per-class spatial attention maps via softmax, and $\mathbf{F}_{\mathrm{fuse}}$ is aggregated over all spatial locations for each class to produce the initial node feature:
\begin{equation}
  \mathbf{H}^{(0)}_{b,k} = \frac{\sum_{h,w} \mathrm{softmax}_k(\mathbf{L}_0)_{b,k,h,w} \cdot \mathbf{F}_{\mathrm{fuse},b,:,h,w}}{\sum_{h,w} \mathrm{softmax}_k(\mathbf{L}_0)_{b,k,h,w} + \epsilon},
\end{equation}
where $\epsilon = 10^{-6}$. Soft attention pooling allows each pixel to contribute to all category nodes weighted by its predicted class distribution, so that ambiguous pixels spread their representations across multiple categories rather than being exclusively assigned to a single potentially incorrect one, mitigating the error propagation inherent in hard assignment. Because the aggregation is image-specific, the resulting node features reflect the actual category composition of the current scene.

\subsubsection{Prior-Biased Graph Reasoning}

Once the nodes are initialized, we refine them with a prior-biased extension of graph attention network (GAT)~\cite{velickovic2018graph} (Fig.~\ref{fig:sgcm}(c)). We inject the prior adjacency $\tilde{\mathbf{A}}$ as an additive bias into the attention logits, so that category pairs with stronger prior affinity receive proportionally larger attention weights:
\begin{equation}
  e_{ij} = \mathrm{LeakyReLU}\!\left(\mathbf{a}^{\top}\!\left[\mathbf{W}\mathbf{h}_i \,\|\, \mathbf{W}\mathbf{h}_j\right]\right),
\end{equation}
\begin{equation}
  \alpha_{ij} = \mathrm{softmax}_{j}\!\left(e_{ij} + \log(\tilde{\mathbf{A}}_{ij} + \epsilon)\right),
\end{equation}
where $e_{ij}$ is the standard GAT attention logit from node $i$ to node $j$, $\mathbf{h}_i$ and $\mathbf{h}_j$ are the current feature vectors of the two nodes, $\mathbf{W}$ is a learnable linear projection matrix, $\mathbf{a}$ is a learnable attention weight vector, $\|$ denotes concatenation, and $\tilde{\mathbf{A}}_{ij}$ is the $(i,j)$-th entry of the fused prior adjacency matrix defined in Eq.~\eqref{eq:prior_adj}. The additive $\log \tilde{\mathbf{A}}_{ij}$ term biases the attention toward category pairs with stronger prior affinity.
Edges with $\tilde{\mathbf{A}}_{ij} < 10^{-5}$ are hard-masked to $-\infty$ before softmax, so that graph reasoning only propagates information along edges with meaningful prior support.

We use a two-layer, 4-head GAT, concatenating the multi-head outputs in the first layer and averaging them in the second. The refined node features $\mathbf{H}_{\mathrm{ref}} \in \mathbb{R}^{B \times K \times D}$ ($D = 512$) serve as context-enhanced category embeddings and are scored against every spatial location of $\mathbf{F}_{\mathrm{fuse}}$ via inner product to produce the graph logits: $\mathbf{L}_{g,b,k,h,w} = \mathbf{H}_{\mathrm{ref},b,k}^{\top} \mathbf{F}_{\mathrm{fuse},b,:,h,w}$. The final prediction fuses the base and graph logits as $\mathbf{L} = \gamma \cdot \mathbf{L}_0 + (1-\gamma) \cdot \mathbf{L}_g$ with $\gamma = 0.85$ (Section~\ref{Ablations}). 
\begin{figure*}[!t]
\centering
\includegraphics[width=.95\linewidth]{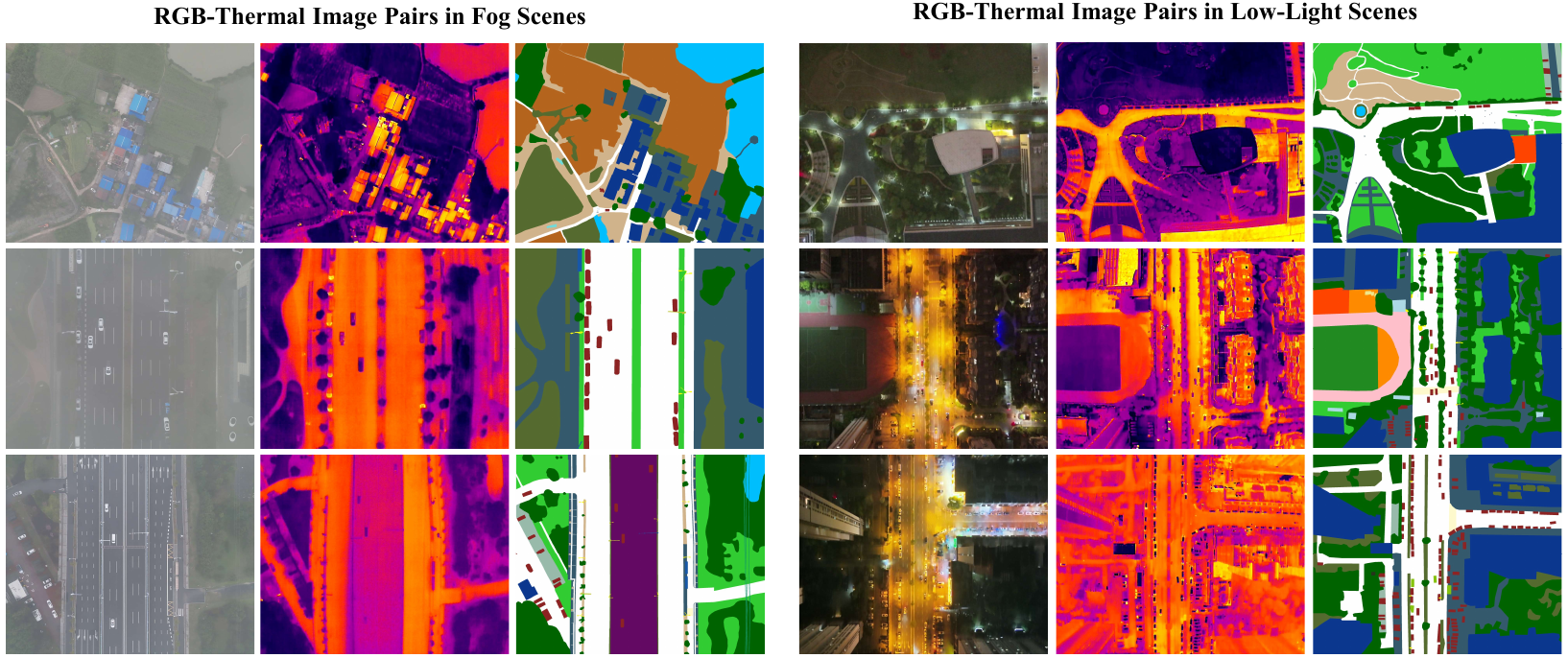}
\caption{RGB, thermal, and semantic annotation examples in URTF. The left side shows groups of RGB images, thermal images, and ground-truth labels captured under cloudy/foggy conditions with increasing fog density. The right side presents groups obtained under low-light conditions, ranging from evening to late night.}
\label{fig:condition}
\end{figure*}

\begin{figure*}[!t]
\centering
\includegraphics[width=\linewidth]{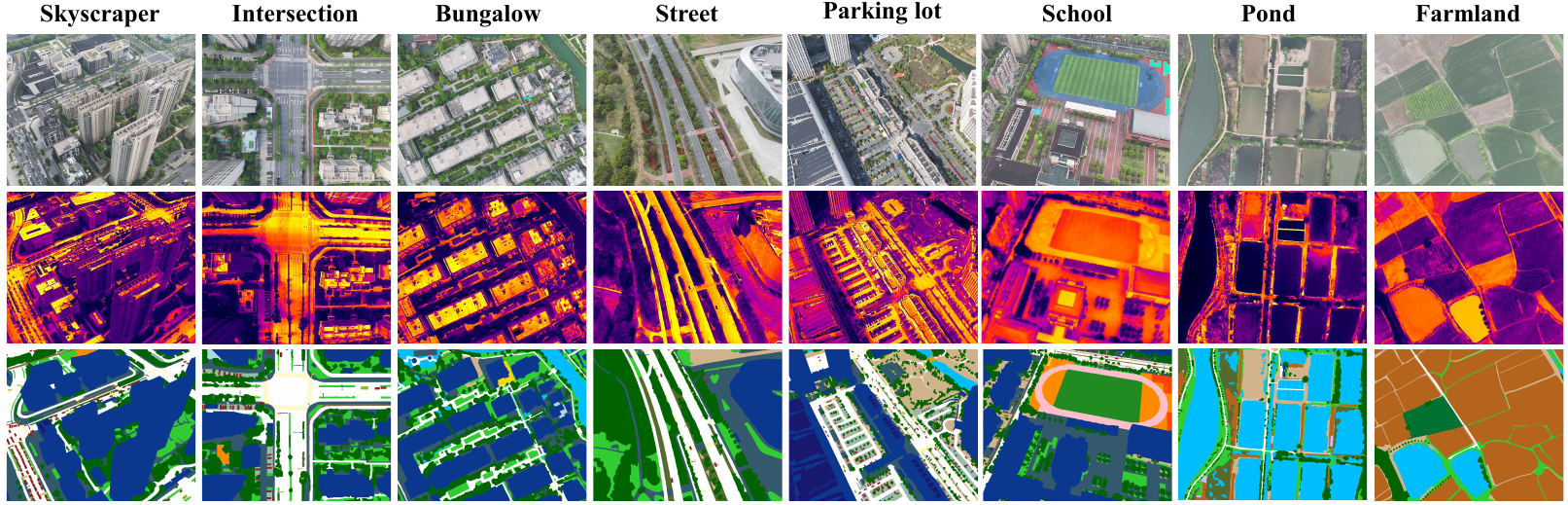}
\caption{Key scenes in the URTF dataset: Skyscraper, Intersection, Bungalow, Street, Parking Lot, School, Pond, and Farmland, all captured at altitudes of 50--300\,m.}
\label{fig:scene}
\end{figure*}

\subsection{Loss Function}
\label{sec:loss}

The network is trained end-to-end with the following objective:
\begin{equation}
  \mathcal{L}_{\mathrm{total}} = \mathcal{L}_{\mathrm{seg}} + \lambda_{\mathrm{dis}} \mathcal{L}_{\mathrm{dis}} + \lambda_{\mathrm{kg}} \mathcal{L}_{\mathrm{kg}},
\end{equation}
where $\mathcal{L}_{\mathrm{seg}}$ is the online hard example mining (OHEM) cross-entropy loss on the final prediction with hard-pixel threshold $\theta = 0.7$; $\mathcal{L}_{\mathrm{kg}} = \|\mathbf{A}_{\delta}\|_1$ is an $\ell_1$ penalty that keeps the learnable residual adjacency sparse; and $\mathcal{L}_{\mathrm{dis}}$ aggregates the three AFD constraints across all four encoder stages:
\begin{equation}
  \mathcal{L}_{\mathrm{dis}} = \frac{1}{4} \sum_{i=1}^{4} \Bigl(
    \lambda_{\mathrm{align}} \mathcal{L}_{\mathrm{align}}^{(i)}
    + \lambda_{\mathrm{sem}} \mathcal{L}_{\mathrm{sem}}^{(i)}
    + \lambda_{\mathrm{orth}} \mathcal{L}_{\mathrm{orth}}^{(i)}
  \Bigr).
\end{equation}
The decoupling weight $\lambda_{\mathrm{dis}} = 0.1$ and the graph fusion weight $\gamma = 0.85$ are selected based on the sensitivity analysis in Section~\ref{Ablations}. The remaining weights are set to $\lambda_{\mathrm{align}} = 0.2$, $\lambda_{\mathrm{sem}} = 0.1$, $\lambda_{\mathrm{orth}} = 0.05$, and $\lambda_{\mathrm{kg}} = 0.01$ throughout all experiments.

\section{URTF Benchmark}

\begin{table*}[t]
\caption{Comparison of URTF with Other Datasets. \textbf{Data}: data composition. \textbf{Reg.}: Registration (Strict = pixel-level; None = unaligned). \textbf{Fine.}: Fine-grained. \textbf{Res.}: resolution. \textbf{\%Anno.}: annotated-pixel ratio. Entries marked with ``--'' are not reported.}
\label{tab:dataset_comparison}
\centering
{\footnotesize
\setlength{\tabcolsep}{3pt}
\renewcommand{\arraystretch}{0.95}
\resizebox{\textwidth}{!}{%
\begin{tabular}{llccccccccccc}
\toprule
\textbf{Category} & \textbf{Dataset} & \textbf{Year} & \textbf{RGB} & \textbf{Thermal} & \textbf{UAV} & \textbf{Data} & \textbf{Reg.} & \textbf{Fine.} & \textbf{\#Imgs} & \textbf{\#Cls} & \textbf{Res.} & \textbf{\%Anno.} \\ \midrule
\multirow{3}{*}{RGB}
 & UAVid~\cite{lyu2020uavid}            & 2020 & \ding{51} & \ding{55} & \ding{51} & Real & -- & \ding{55} & 420     & 8  & 3840$\times$2160 & 82.69 \\
 & FloodNet~\cite{rahnemoonfar2021floodnet} & 2021 & \ding{51} & \ding{55} & \ding{51} & Real & -- & \ding{55} & 2,343   & 10 & 4000$\times$3000 & -- \\
 & VDD~\cite{cai2025vdd}                & 2025 & \ding{51} & \ding{55} & \ding{51} & Real & -- & \ding{55} & 400     & 7  & 4000$\times$3000 & -- \\ \midrule
\multirow{10}{*}{RGB-T}
 & MFNet~\cite{ha2017mfnet}              & 2017 & \ding{51} & \ding{51} & \ding{55} & Real & Strict & \ding{55} & 1,569   & 9  & 640$\times$480   & 7.86 \\
 & PST900~\cite{shivakumar2020pst900}    & 2020 & \ding{51} & \ding{51} & \ding{55} & Real & Strict & \ding{55} & 894     & 5  & 1280$\times$720  & 3.02 \\
 & SemanticRT~\cite{ji2023semanticrt}  & 2023 & \ding{51} & \ding{51} & \ding{55} & Real & Strict & \ding{55} & 11,371  & 13 & 1280$\times$1024 & 21.27 \\
 & FMB~\cite{liu2023multi}               & 2023 & \ding{51} & \ding{51} & \ding{55} & Real & Strict & \ding{55} & 1,500   & 15 & 800$\times$600   & 98.16 \\
 & CART~\cite{chen2024cart}              & 2024 & \ding{51} & \ding{51} & \ding{51} & Real & Strict & \ding{55} & 2,282   & 11 & 960$\times$600   & 99.98 \\
 & MVSeg~\cite{zhang2024mrfs}            & 2024 & \ding{51} & \ding{51} & \ding{55} & Real & Strict & \ding{55} & 3,545   & 26 & 480$\times$640   & 98.96 \\
 & MVUAV~\cite{ji2024unleashing}         & 2024 & \ding{51} & \ding{51} & \ding{51} & Real & Strict & \ding{55} & 2,183   & 36 & 1920$\times$1080 & 99.18 \\ 
 & Kust4K~\cite{ouyang2025kust4k}        & 2025 & \ding{51} & \ding{51} & \ding{51} & Real & Strict & \ding{55} & 4,024   & 8  & 640$\times$512   & 77.34 \\
 & U-MFNet~\cite{zhou2025drgbt}           & 2025 & \ding{51} & \ding{51} & \ding{55} & Synth. & None & \ding{55} & 1,569   & 9  & 640$\times$480   & 7.86 \\
\rowcolor{gray!15}
 & \textbf{URTF (Ours)}                  & \textbf{--} & \textbf{\ding{51}} & \textbf{\ding{51}} & \textbf{\ding{51}} & \textbf{Real+Synth.} & \textbf{None} & \textbf{\ding{51}} & \textbf{25,519} & \textbf{61} & \textbf{640$\times$512} & \textbf{99.992} \\ \bottomrule
\end{tabular}
}
}
\end{table*}

\begin{figure}[t]
  \centering
  \includegraphics[width=\linewidth]{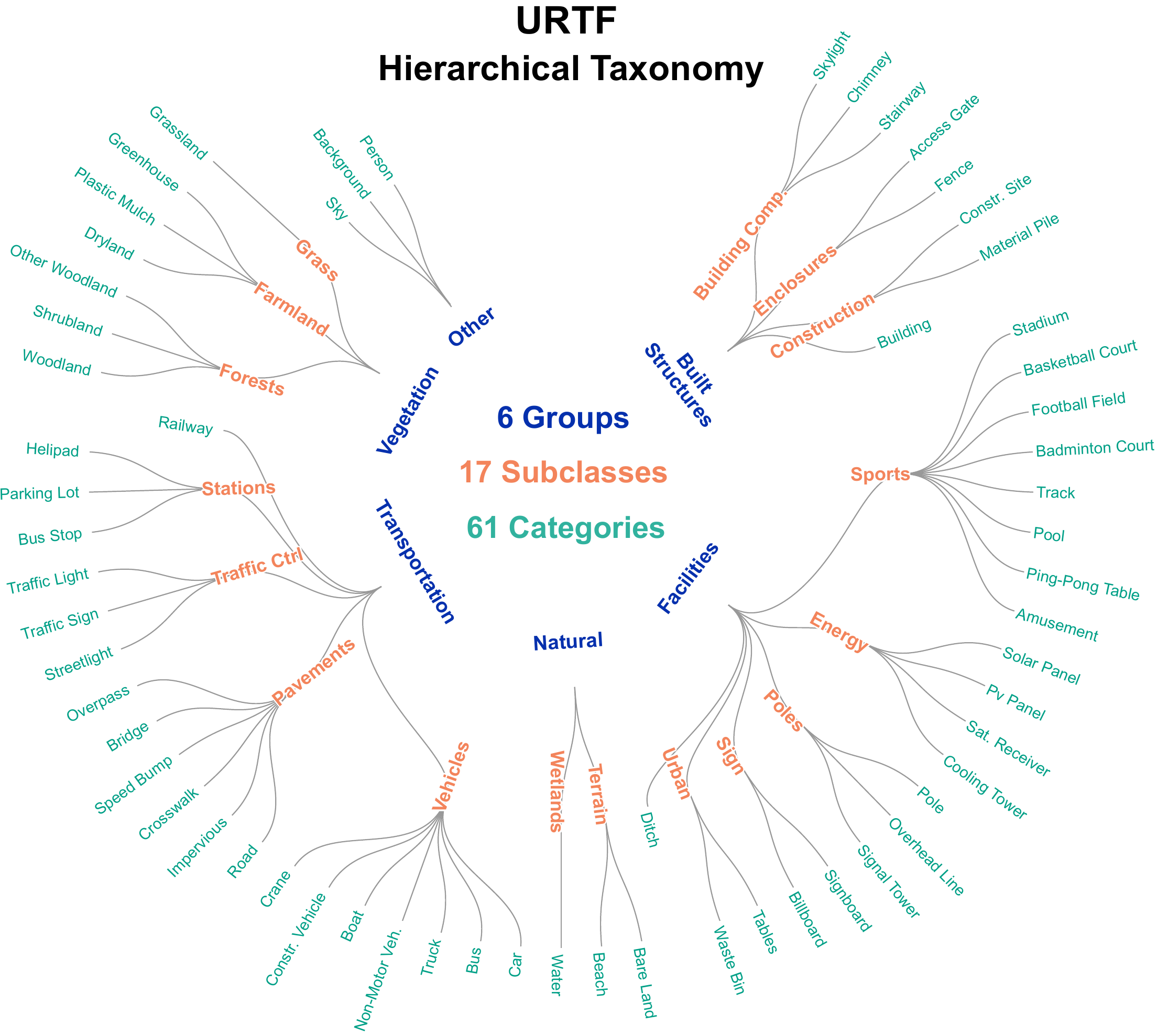}
\caption{Hierarchical taxonomy of semantic categories in URTF. Blue, orange, and green labels denote top-level groups, intermediate subclasses, and leaf categories, respectively.}
  \label{fig:dataset1}
\end{figure}

\subsection{Comparison with Existing Datasets}

Table~\ref{tab:dataset_comparison} summarizes existing RGB-Thermal benchmarks. Most were collected from ground-level platforms~\cite{ha2017mfnet,shivakumar2020pst900,liu2023multi,vertens2020heatnet}, and the few UAV-oriented datasets remain limited in scale or annotation granularity~\cite{ji2024unleashing,ouyang2025kust4k}. Recent efforts such as MVUAV~\cite{ji2024unleashing} have extended the label space to 36 categories and introduced several object subtypes, representing a notable step toward finer annotation. However, such granularity remains confined to a few object families rather than a broad, taxonomy-level design, and the dataset still relies on strict pixel-level registration. In contrast, URTF provides 61 semantic categories organized in a three-level hierarchy across six top-level groups (Fig.~\ref{fig:dataset1}) and preserves the realistic cross-modal misalignment produced by practical dual-sensor UAV hardware, constituting the largest and most comprehensive fine-grained benchmark for unaligned UAV RGBT semantic segmentation available to date.

\subsection{Dataset Construction}

\begin{figure}[t]
  \centering
  \includegraphics[width=\linewidth]{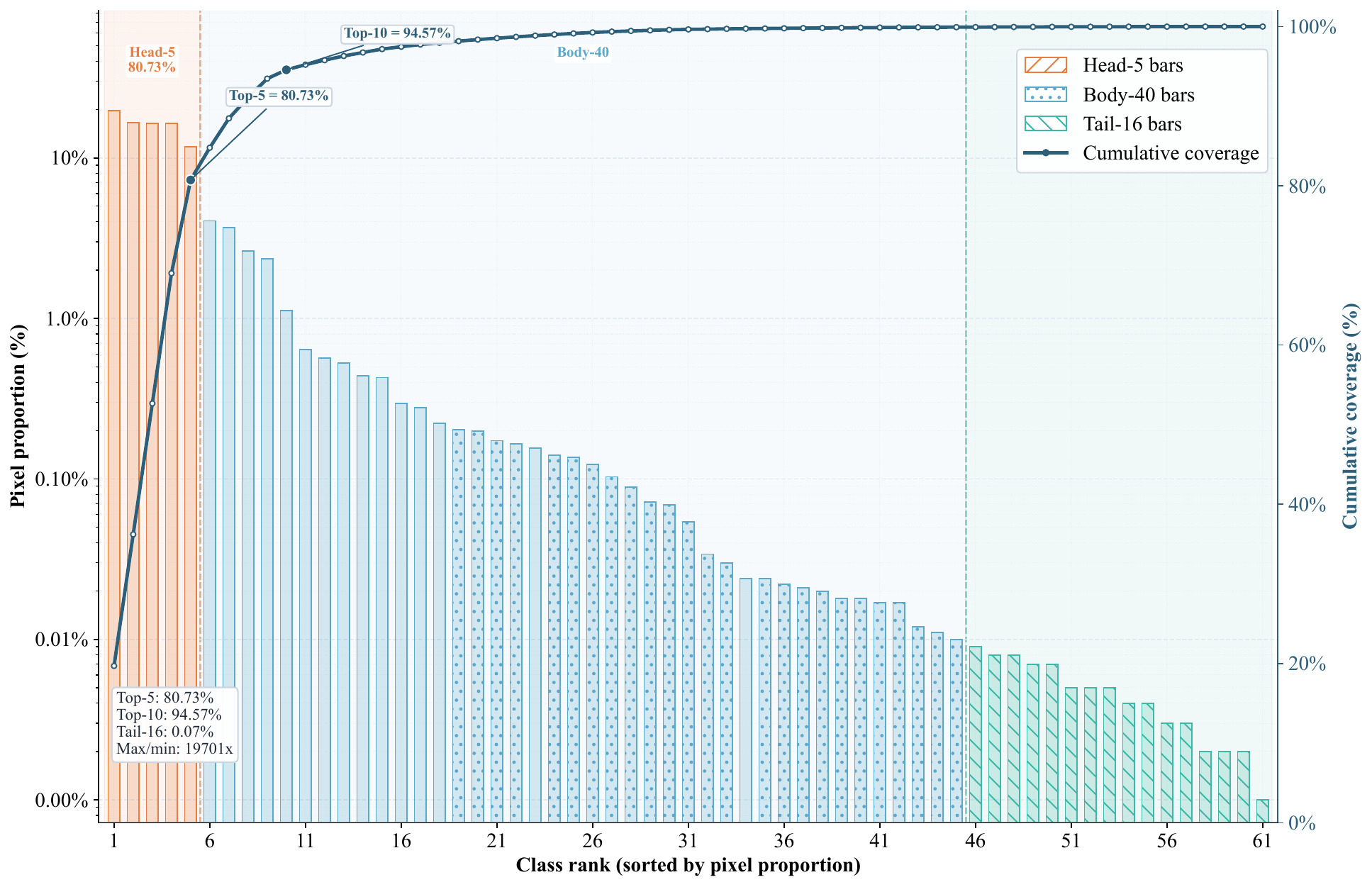}
\caption{Long-tailed pixel distribution of the 61 semantic categories in URTF. Categories are sorted by pixel proportion, with the Head-5, Body-40, and Tail-16 partitions highlighted together with the cumulative pixel coverage.}
  \label{fig:dataset2}
\end{figure}

We collect URTF with DJI M30T and DJI Matrice 3TD drones at altitudes of 50--300\,m over urban, suburban, farmland, and water-body scenes (Fig.~\ref{fig:scene}), covering daytime, foggy, and low-light conditions (Fig.~\ref{fig:condition}). Each drone carries a dual-sensor gimbal with a high-resolution RGB camera and a $640\times512$ thermal camera. 
The two sensors have different optical centers, focal lengths, and imaging resolutions, so their images share an approximate field of view but lack exact pixel correspondence. Because the relative displacement between the two sensors varies with flight altitude, object depth, and platform motion, a calibration matrix estimated on one scene cannot remove local offsets in another, especially near object boundaries or elevated structures, and global registration would introduce interpolation artifacts and distort thin objects. We therefore only resize and crop RGB frames to $640\times512$ to match the thermal resolution, without any manual fine alignment, automated registration, or post-hoc correction. Throughout this paper, \emph{unaligned} means the two modalities cover approximately the same scene area without pixel-to-pixel geometric correspondence.

URTF defines 61 semantic categories organized into a 3-level taxonomy with 6 top-level groups (Fig.~\ref{fig:dataset1}), resulting in a strongly long-tailed distribution where the top-5 categories account for over 80\% of pixels and the bottom-16 each occupy less than 0.01\% (Fig.~\ref{fig:dataset2}). We annotate the modality that provides clearer object boundaries under each imaging condition (RGB for daytime, thermal for nighttime), so the ground-truth labels reside in the reference frame of the annotated modality.  An internal pre-trained segmentation model generates initial masks, which are then refined by \textbf{30 professional annotators} over \textbf{6 months} (\textbf{8 hours per day}, totaling more than \textbf{28,000 person-hours}). Every image pair passes through \textbf{three independent rounds of expert quality inspection}; inaccurate or ambiguous labels are returned for correction before advancing to the next round, yielding a final annotated-pixel coverage of \textbf{99.992\%}. To further increase scene diversity and mitigate the long-tailed distribution, we supplement the real data with 8\,625 synthetic image pairs rendered in AirSim and CARLA, enriching rare scene configurations and under-represented tail categories.

\begin{table*}[!t]
\caption{Quantitative Comparison with State-of-the-Art Methods on URTF. \textbf{mIoU}: mean Intersection over Union across all 61 categories. \textbf{mAcc}: mean per-class accuracy. \textbf{aAcc}: overall pixel accuracy. \textbf{Head-5}: mean IoU of the 5 dominant categories that collectively account for 80.73\% of all pixels. \textbf{Tail-16}: mean IoU of the 16 rarest categories, each occupying less than 0.01\% of pixels. Methods listed under \textit{RGB-T Fusion} use paired RGB-T inputs. \textbf{\textcolor{red}{Bold red}}: best; \textbf{\textcolor{blue}{bold blue}}: second best.}
\label{tab:sota_comparison}
\centering
\setlength{\tabcolsep}{7pt}
\renewcommand{\arraystretch}{1.08}
\resizebox{0.95\textwidth}{!}{
\begin{tabular}{cccccccc}
\toprule
\textbf{Method} & \textbf{Pub.} & \textbf{Modality}
  & \textbf{mIoU (\%)} & \textbf{mAcc (\%)} & \textbf{aAcc (\%)}
  & \textbf{Head-5 (\%)} & \textbf{Tail-16 (\%)} \\
\midrule
\multicolumn{8}{l}{\textit{RGB-only Methods}} \\
DeepLabv3+~\cite{chen2018deeplabv3plus} & ECCV '18    & RGB   & 53.63 & 66.05 & 88.51 & 81.53 & 33.35 \\
HRNet~\cite{sun2019hrnet}               & CVPR '19    & RGB   & 56.63 & 65.66 & 88.92 & 81.73 & 40.70 \\
SegFormer~\cite{xie2021segformer}       & NeurIPS '21 & RGB   & 60.05 & 69.53 & 89.58 & 82.72 & 45.79 \\
Mask2Former~\cite{cheng2021mask2former} & CVPR '22    & RGB   & 59.64 & 71.56 & 88.44 & 81.49 & 45.28 \\
SegNeXt~\cite{guo2022segnext}           & NeurIPS '22 & RGB   & 59.22 & 70.22 & 89.34 & 82.33 & 44.36 \\
EFENet~\cite{Chen2024EdgeFE}            & TGRS '24    & RGB   & 59.83 & 70.71 & 89.08 & 81.94 & 46.97 \\
\midrule
\multicolumn{8}{l}{\textit{Thermal-only Methods}} \\
DeepLabv3+~\cite{chen2018deeplabv3plus} & ECCV '18    & Thermal & 24.45 & 31.96 & 75.59 & 63.85 &  7.69 \\
HRNet~\cite{sun2019hrnet}               & CVPR '19    & Thermal & 27.87 & 33.23 & 78.44 & 66.65 & 11.44 \\
SegFormer~\cite{xie2021segformer}       & NeurIPS '21 & Thermal & 34.95 & 42.29 & 80.32 & 69.26 & 18.94 \\
Mask2Former~\cite{cheng2021mask2former} & CVPR '22    & Thermal & 28.54 & 36.99 & 77.99 & 67.48 &  6.99 \\
SegNeXt~\cite{guo2022segnext}           & NeurIPS '22 & Thermal & 32.81 & 39.08 & 80.26 & 69.32 & 16.51 \\
EFENet~\cite{Chen2024EdgeFE}            & TGRS '24    & Thermal & 33.00 & 40.12 & 80.01 & 68.64 & 18.69 \\
\midrule
\multicolumn{8}{l}{\textit{RGB-T Fusion Methods}} \\
EGFNet~\cite{zhou2022edge}                    & AAAI '22   & RGB-T & 48.74 & 52.53 & 84.21 & 74.59 & 18.59 \\
CMX~\cite{zhang2023cmx}                       & TITS '23   & RGB-T & 60.31 & 69.19 & 90.80 & 84.52 & 45.58 \\
CMNext~\cite{zhang2023delivering}             & CVPR '23   & RGB-T & 62.78 & 75.18 & 90.78 & 84.58 & 49.67 \\
SGFNet~\cite{Wang2023sgfnet}                  & TCSVT '23  & RGB-T & 62.41 & 75.38 & 91.08 & 82.97 & 48.22 \\
CRM~\cite{Shin2024ComplementaryRM}            & ICRA '24   & RGB-T & 67.78 & \textbf{\textcolor{blue}{77.01}} & 91.91 & 86.78 & 55.46 \\
GeminiFusion~\cite{jia2024geminifusion}       & ICML '24   & RGB-T & 61.73 & 71.46 & 90.63 & 84.34 & 46.89 \\
MRFS~\cite{zhang2024mrfs}                     & CVPR '24   & RGB-T & 65.28 & 75.13 & 91.29 & 84.21 & 53.87 \\
ASANet~\cite{ZHANG2024574asa}                 & ISPRS '24  & RGB-X & 53.91 & 63.30 & 90.78 & 84.01 & 25.53 \\
MiLNet~\cite{milnet2025tip}                   & TIP '25    & RGB-T & 61.29 & 68.97 & 91.12 & 85.08 & 41.88 \\
DFormerV2~\cite{yin2025dformerv2}             & CVPR '25   & RGB-X & 64.94 & 74.91 & 91.23 & 84.96 & 53.73 \\
AMDANet~\cite{Zhong_2025_ICCV}                & ICCV '25   & RGB-T & 68.13 & 75.46 & 91.49 & 85.37 & 54.77 \\
Mul-VMamba~\cite{mulvmamba2026kbs}            & KBS '26    & RGB-T & 64.92 & 74.03 & 91.34 & 85.33 & 52.47 \\
MambaSeg~\cite{wang2026mambaseg}              & AAAI '26   & RGB-X & \textbf{\textcolor{blue}{68.31}} & 76.62 & \textbf{\textcolor{blue}{92.47}} & \textbf{\textcolor{blue}{87.28}} & \textbf{\textcolor{blue}{55.54}} \\
\midrule
\textbf{GSCNet (Ours)} & \textbf{--} & \textbf{RGB-T}
  & \textbf{\textcolor{red}{71.04}} & \textbf{\textcolor{red}{78.97}} & \textbf{\textcolor{red}{92.65}} & \textbf{\textcolor{red}{87.40}} & \textbf{\textcolor{red}{60.17}} \\
\bottomrule
\end{tabular}
}
\end{table*}

\subsection{Dataset Statistics and Core Challenges}

URTF contains 25,519 RGB-Thermal image pairs, including 16,894 real samples and 8,625 synthetic ones. We use 20,393 pairs for training and 5,126 for validation, roughly a 4:1 split, and keep the validation set entirely real-world. In terms of illumination and weather conditions, all 25,519 image pairs are distributed across three distinct scenarios: 14,844 daytime, 6,900 foggy, and 3,775 low-light, covering the diverse sensing conditions encountered in practical UAV deployments.

The benchmark is challenging for three reasons. First, cross-modal spatial misalignment caused by sensor parallax and vibration cannot be removed by a single global transform. Second, the class distribution is severely long-tailed: the top-5 categories already occupy 80.73\% of all pixels, whereas 51 of the 61 classes each account for less than 1\% (Fig.~\ref{fig:dataset2}). The five head categories (Road, Woodland, Building, Grassland, Impervious) together occupy 80.73\% of pixels, while the 16 tail categories (e.g., Crane, Helipad, Cooling Tower, Greenhouse) each contribute less than 0.01\%. Third, many categories remain difficult to distinguish from the UAV viewpoint, including Woodland/Shrubland, Road/Impervious, Pole/Streetlight/Traffic Light, and several vehicle subtypes.

\section{Experiments}

\begin{figure*}[!t]
    \centering
    \includegraphics[width=\linewidth]{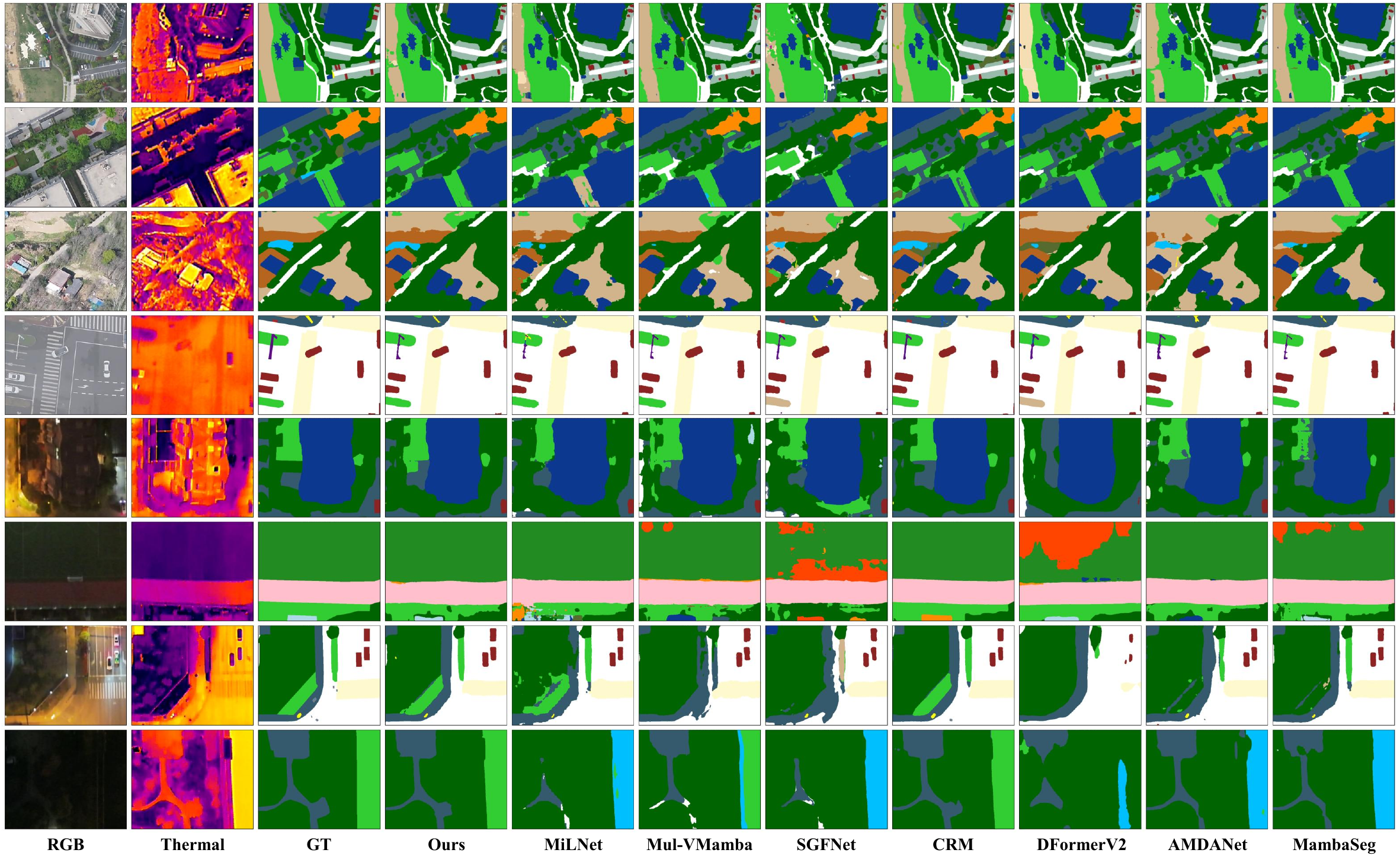}
    \caption{Visual comparisons among GSCNet and seven competing methods on the URTF benchmark in daytime (first 4 rows) and nighttime (last 4 rows) scenes.}
    \label{fig:sota_compar}
\end{figure*}

We evaluate GSCNet against state-of-the-art methods on the URTF validation set under a unified protocol without test-time augmentation.

\subsection{Implementation Details}
\label{Implementation}

GSCNet uses MiT-B4~\cite{xie2021segformer} (ImageNet-1K pre-trained) as the backbone and is implemented in PyTorch on a single NVIDIA H20 GPU. All methods, including GSCNet, are trained for 50 epochs with batch size 12, AdamW optimizer (lr $1 \times 10^{-4}$, weight decay $0.01$), poly LR schedule with 5-epoch warmup, an input resolution of $640\times512$, and identical augmentation (random crop, horizontal flip, multi-scale resizing $\{0.5$--$1.75\}$, photometric distortion). GSCNet-specific loss weights are $\lambda_{\mathrm{dis}}=0.1$, $\lambda_{\mathrm{kg}}=0.01$. Each comparison method retains its official backbone; the three RGB-X methods (ASANet, DFormerV2, MambaSeg) only modify the input layer to accept RGB-Thermal input. No test-time augmentation or cross-modal pre-alignment is applied, preserving realistic sensor offsets.

\subsection{Comparison with State-of-the-Art Methods}
\label{Quantitative}
 
We compare GSCNet with 19 methods across 25 configurations spanning RGB-only, thermal-only, RGB-T fusion, and RGB-X cross-modal settings, with the full results reported in Table~\ref{tab:sota_comparison}. The three RGB-X models, ASANet, DFormerV2, and MambaSeg, were originally developed for RGB-SAR, RGB-D, and RGB-event segmentation and are retrained on URTF under the same protocol as cross-modal generalization baselines. MRFS~\cite{zhang2024mrfs} and AMDANet~\cite{Zhong_2025_ICCV} adopt a dual-task design that jointly optimizes image fusion and semantic segmentation; we retain their original architecture but only report segmentation metrics, as URTF does not provide image-fusion ground truth.

GSCNet reaches 71.04\% mIoU on URTF, exceeding MambaSeg by 2.73\%, AMDANet by 2.91\%, and SegFormer (strongest RGB-only baseline) by 10.99\%. The gap widens on tail categories: GSCNet obtains 60.17\% Tail-16 IoU versus 55.54\% for MambaSeg and 54.77\% for AMDANet. Models without misalignment handling, such as EGFNet (48.74\%), fail to exploit the thermal modality, and even the best competing fusion methods trail GSCNet by 4.63--5.40\% on Tail-16, showing that fusion alone cannot resolve fine-grained tail categories without dedicated alignment and semantic reasoning.

Adding thermal input does not guarantee better segmentation on URTF. Several RGB-T fusion models score below the RGB-only SegFormer baseline. EGFNet and ASANet rely on local consistency between modalities for boundary-level or cross-modal fusion, so misaligned thermal features actively hurt their decoders. Stronger models such as AMDANet and MambaSeg use more flexible cross-modal interaction and partially mitigate this problem, but they still lack explicit mechanisms for local geometric correction and category-level calibration. GSCNet improves both mIoU and Tail-16 because FDAM first reduces feature-level misalignment and SGCM then corrects class-level confusion among rare or visually similar categories.

Fig.~\ref{fig:sota_compar} shows qualitative results. In daytime scenes (rows~1--4), RGB provides clear contours, but thermal responses shift around object boundaries and create duplicate activations after naive fusion. Competing methods therefore produce ghosting artifacts and blurred boundaries. The foggy scene in row~4 amplifies this problem for small targets such as traffic lights and streetlights. GSCNet avoids these artifacts because FDAM corrects feature-level misalignment before fusion. In nighttime scenes (rows~5--8), thermal cues become dominant while RGB textures turn noisy or vanish. Methods without adaptive reference selection tend to overfit the RGB modality and suffer large-area failures and category mislabeling. GSCNet remains stable because IAA selects the more reliable modality under each imaging condition and SGCM suppresses contextually unlikely labels.

Table~\ref{tab:efficiency} reports model complexity and inference speed. Compared with MambaSeg, GSCNet nearly halves the parameter count (160.66M vs.\ 315.44M) while improving mIoU from 68.31\% to 71.04\%. Relative to AMDANet, it uses comparable FLOPs (208.07G vs.\ 213.43G) but runs substantially faster (16.62 vs.\ 11.03 FPS). CRM is both larger and slower, yet remains 3.26\% lower in mIoU. The main computational cost comes from the dual MiT-B4 encoders and multi-stage FDAM blocks, whereas SGCM adds only marginal overhead because graph reasoning operates on 61 category nodes rather than dense $H \times W$ spatial grids.
\subsection{Ablation Studies}
\label{Ablations}

\begin{table}[!t]
\caption{Comparison of Model Complexity and Inference Speed. FPS is measured on a single NVIDIA H20 GPU at $640\times512$ resolution with batch size~1. \textbf{Bold} highlights the best result per column.}
\label{tab:efficiency}
\centering
\setlength{\tabcolsep}{3pt} 
\renewcommand{\arraystretch}{1.08}
\resizebox{\columnwidth}{!}{
\begin{tabular}{c|ccccc}
\toprule
\textbf{Method} & \textbf{Backbone} & \textbf{Params/M} & \textbf{FLOPs/G} & \textbf{FPS} & \textbf{mIoU} \\ \midrule
\multicolumn{6}{l}{\textit{RGB-only Methods}} \\
DeepLabv3+~\cite{chen2018deeplabv3plus} & Res101  & 60.24  & 318.00 & 44.49 & 53.63 \\
HRNet~\cite{sun2019hrnet}               & HRNet-W48   & 65.89  & 117.80 & 51.75 & 56.63 \\
SegFormer~\cite{xie2021segformer}       & MiT-B4      & 61.39  &  76.18 & 44.68 & 60.05 \\
Mask2Former~\cite{cheng2021mask2former} & Swin-B      & 107.00 & 172.00 & 24.80 & 59.64 \\
SegNeXt~\cite{guo2022segnext}           & MSCAN-L     & \textbf{48.84} &  82.19 & \textbf{53.15} & 59.22 \\
EFENet~\cite{Chen2024EdgeFE}            & Twins-L     & 143.84 & 198.99 & 22.01 & 59.83 \\ \midrule
\multicolumn{6}{l}{\textit{RGB-T Fusion Methods}} \\
EGFNet~\cite{zhou2022edge}              & Res152        &  62.93 & 230.14 & 19.62 & 48.74 \\
CMX~\cite{zhang2023cmx}                 & MiT-B4           & 143.38 & 172.27 & 22.80 & 60.31 \\
CMNext~\cite{zhang2023delivering}       & MiT-B4           & 116.59 & 154.89 & 24.90 & 62.78 \\
SGFNet~\cite{Wang2023sgfnet}            & Res50         & 125.31 & 157.25 & 24.04 & 62.41 \\
CRM~\cite{Shin2024ComplementaryRM}      & MiT-B4           & 193.51 & 267.48 & 15.60 & 67.78 \\
GeminiFusion~\cite{jia2024geminifusion} & MiT-B4           & 103.32 & 271.75 & 14.20 & 61.73 \\
MRFS~\cite{zhang2024mrfs}               & MiT-B4           & 134.99 & 140.65 & 19.55 & 65.28 \\
ASANet~\cite{ZHANG2024574asa}           & ConvNeXt-T      &  82.93 & 129.80 & 43.20 & 53.91 \\
MiLNet~\cite{milnet2025tip}             & MiT-B4     & 126.04 & 197.33 & 22.40 & 61.29 \\
DFormerV2~\cite{yin2025dformerv2}       & DFormerV2-L           &  95.57 & 146.88 & 17.53 & 64.94 \\
AMDANet~\cite{Zhong_2025_ICCV}          & MiT-B4           & 135.76 & 213.43 & 11.03 & 68.13 \\
Mul-VMamba~\cite{mulvmamba2026kbs}      & VMamba-T          & 112.23 & \textbf{56.33} & 40.40 & 64.92 \\
MambaSeg~\cite{wang2026mambaseg}        & VMamba-T          & 315.44 & 186.06 & 20.92 & 68.31 \\
\midrule
\textbf{GSCNet (Ours)}                  & MiT-B4   & 160.66 & 208.07 & 16.62 & \textbf{71.04} \\ \bottomrule
\end{tabular}
}
\end{table}

We analyze FDAM and SGCM under the same training setting as Section~\ref{Implementation}. Tables~\ref{tab:ablation_overview}--\ref{tab:ablation_sgcm} report individual and joint effects, and Fig.~\ref{fig:hyperparam} examines the two key hyperparameters. The ablation baseline is derived from AMDANet~\cite{Zhong_2025_ICCV} after removing the image fusion head and replacing its specialized fusion module with lightweight channel-wise concatenation, yielding 67.63\% mIoU and 52.67\% Tail-16 on URTF.

\subsubsection{Module-Level Overview}

FDAM and SGCM target different error sources. FDAM improves mIoU by 2.50\% and SGCM by 2.67\%, confirming that spatial misalignment and semantic confusion are both significant bottlenecks. Combining them raises the gain to 3.41\%, less than the arithmetic sum, which indicates partial overlap but clear complementarity.

The Tail-16 numbers clarify the roles of the two modules. FDAM raises Tail-16 from 52.67\% to 57.23\%, showing that rare categories also benefit from cleaner cross-modal boundaries. SGCM raises Tail-16 to 58.75\%, a larger tail gain than its mIoU gain, consistent with its design goal of using class relations to support rare labels. The full model reaches 60.17\%, indicating that rare-category recognition needs both reliable spatial evidence and relational semantic context.

\begin{table}[!t]
\caption{Module-Level Ablation on the URTF Validation Set. Gains are relative to the baseline.}
\label{tab:ablation_overview}
\centering
\resizebox{\columnwidth}{!}{
\begin{tabular}{cc|cc|c}
\toprule
\textbf{FDAM} & \textbf{SGCM} & \textbf{mIoU (\%)} & \textbf{Tail-16 (\%)} & \textbf{$\Delta$mIoU} \\ \midrule
             &              & 67.63 & 52.67 & -- \\
\checkmark   &              & 70.13 & 57.23 & +2.50 \\
             & \checkmark   & 70.30 & 58.75 & +2.67 \\
\checkmark   & \checkmark   & 71.04 & 60.17 & +3.41 \\ \bottomrule
\end{tabular}
}
\end{table}

\subsubsection{FDAM Internal Ablation}

We next examine which part of FDAM contributes most. AFD introduces an asymmetric encoder and three loss constraints ($\mathcal{L}_{\mathrm{align}}$, $\mathcal{L}_{\mathrm{sem}}$, $\mathcal{L}_{\mathrm{orth}}$) to separate shared structural cues from modality-private perceptual cues, while IAA estimates deformable offsets in the shared structural space.
Most of the gain comes from disentangling structure before alignment. The asymmetric encoder alone brings a 1.21\% mIoU gain, and adding the decoupling losses increases it to 1.72\%, showing that a cleaner shared subspace is crucial for robust fusion. IAA adds a further 0.78\% once the shared representation has been stabilized. The ordering matters: decoupling must precede deformable alignment, because offset estimation in the raw feature space confuses appearance gaps with spatial displacements (Section~\ref{sec:fdam}).

\begin{table}[!t]
\caption{FDAM Internal Ablation on the URTF Validation Set. SGCM is disabled throughout. \textbf{Asym.}: Asymmetric encoder. \textbf{$\mathcal{L}_{\mathrm{dis}}$}: Three decoupling loss constraints. \textbf{IAA}: Illumination-aware alignment.}
\label{tab:ablation_fdam}
\centering
\resizebox{\columnwidth}{!}{
\begin{tabular}{ccc|cc|c}
\toprule
\textbf{Asym.} & \textbf{$\mathcal{L}_{\mathrm{dis}}$} & \textbf{IAA} & \textbf{mIoU (\%)} & \textbf{Tail-16 (\%)} & \textbf{$\Delta$mIoU} \\ \midrule
             &              &              & 67.63 & 52.67 & --    \\
\checkmark   &              &              & 68.84 & 54.88 & +1.21 \\
\checkmark   & \checkmark   &              & 69.35 & 56.12 & +1.72 \\
\checkmark   & \checkmark   & \checkmark   & \textbf{70.13} & \textbf{57.23} & \textbf{+2.50} \\ \bottomrule
\end{tabular}
}
\end{table}

\begin{table}[!t]
\caption{SGCM Knowledge Source Ablation on the URTF Validation Set. FDAM is disabled throughout. \textbf{$\mathbf{A}_H$}: Taxonomic hierarchy prior. \textbf{$\mathbf{A}_C$}: Co-occurrence prior. \textbf{$\mathbf{A}_{\delta}$}: Learnable residual adjacency.}
\label{tab:ablation_sgcm}
\centering
\resizebox{\columnwidth}{!}{
\begin{tabular}{ccc|cc|c}
\toprule
\textbf{$\mathbf{A}_H$} & \textbf{$\mathbf{A}_C$} & \textbf{$\mathbf{A}_{\delta}$} & \textbf{mIoU (\%)} & \textbf{Tail-16 (\%)} & \textbf{$\Delta$mIoU} \\ \midrule
             &              &              & 67.63 & 52.67 & --    \\
\checkmark   &              &              & 68.88 & 55.93 & +1.25 \\
             & \checkmark   &              & 69.08 & 56.35 & +1.45 \\
\checkmark   & \checkmark   &              & 69.81 & 57.76 & +2.18 \\
\checkmark   & \checkmark   & \checkmark   & \textbf{70.30} & \textbf{58.75} & \textbf{+2.67} \\ \bottomrule
\end{tabular}
}
\end{table}
\subsubsection{SGCM Knowledge Source Ablation}

\begin{figure}[!t]
    \centering
    \includegraphics[width=\linewidth]{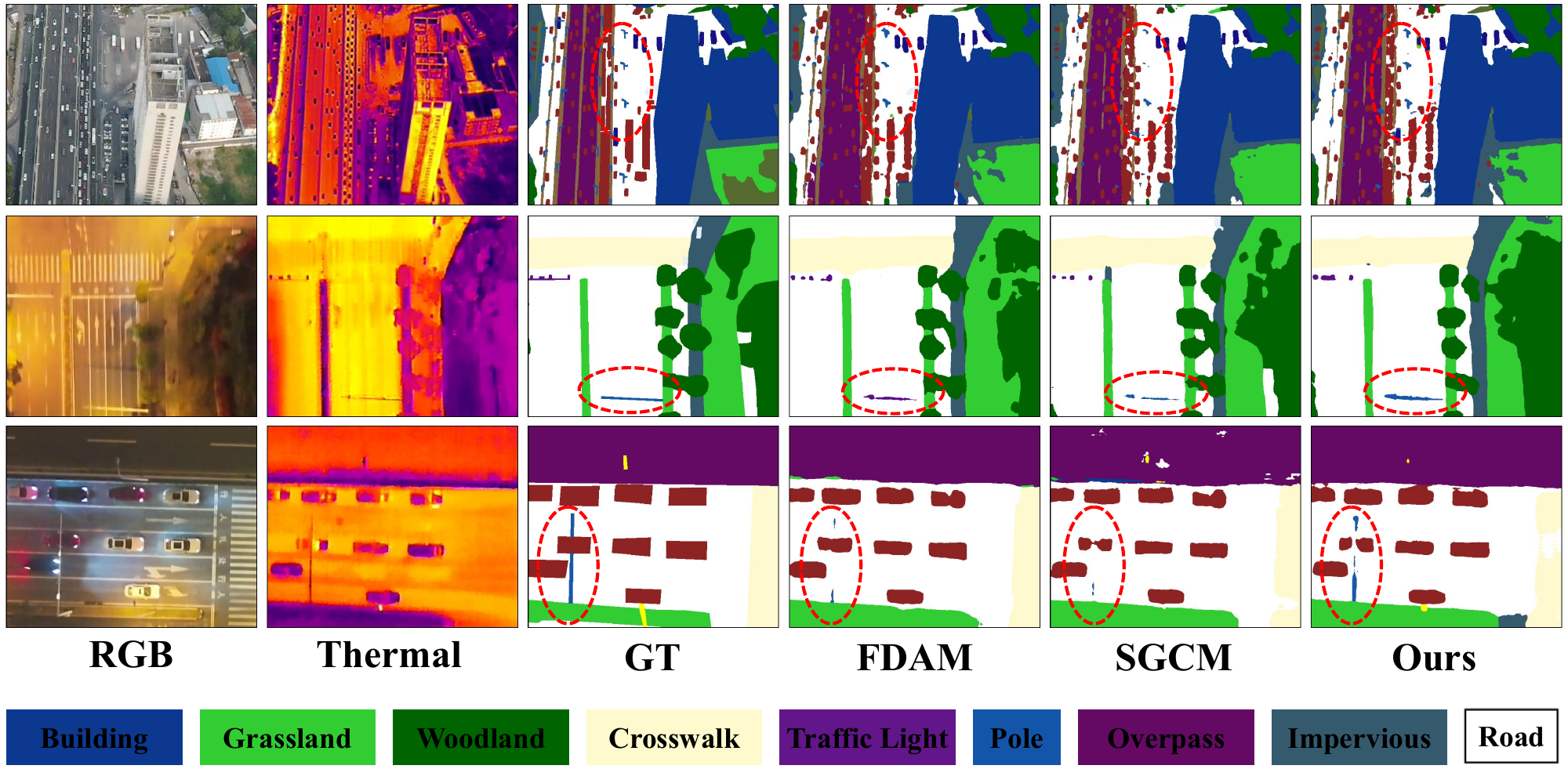}
    \caption{Qualitative ablation on the URTF benchmark. From left to right: RGB image, thermal image, ground truth, FDAM-only, SGCM-only, and full GSCNet. }
    \label{fig:ablation_compar}
\end{figure}

We then examine each knowledge source in SGCM. The adjacency matrix combines a taxonomic hierarchy prior $\mathbf{A}_H$, a co-occurrence prior $\mathbf{A}_C$, and a learnable residual $\mathbf{A}_{\delta}$.
Both static priors help on their own, with $\mathbf{A}_C$ slightly stronger than $\mathbf{A}_H$; combining them raises performance to 69.81\% mIoU and 57.76\% Tail-16. Taxonomy and scene co-occurrence capture different information: $\mathbf{A}_H$ connects semantically related classes, while $\mathbf{A}_C$ reflects image-level context. Adding $\mathbf{A}_{\delta}$ further improves Tail-16 by 0.99\%, showing that the training data contains useful inter-class relations beyond the fixed priors.
Fig.~\ref{fig:ablation_compar} illustrates complementary failure modes. In row~1, the circled region contains three poles: FDAM segments them correctly through spatial alignment, but SGCM produces fragmented predictions because misalignment is harmful to small, thin targets. In row~2, the circled pole is misclassified as a traffic light by FDAM, while SGCM assigns the correct category but introduces boundary fragmentation without alignment. In row~3, a pole occluding a vehicle challenges both modules individually. Combining FDAM and SGCM yields the best results across all three cases, though residual fragmentation on thin occluding structures remains.
                                                                                                                                                                                                        

\subsubsection{Sensitivity Analysis of Hyperparameters}

\begin{figure}[t]
\centering
\includegraphics[width=\columnwidth]{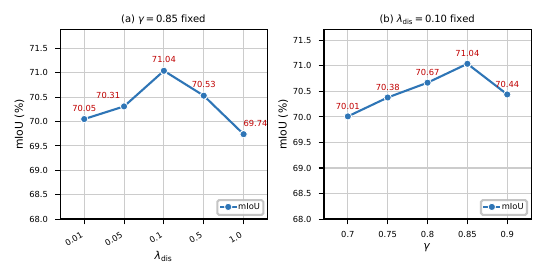}
\caption{Sensitivity analysis of key hyperparameters on the URTF validation set. Red labels show mIoU (\%) at each setting. (a) Varying decoupling weight $\lambda_{\mathrm{dis}}$ with graph fusion weight $\gamma=0.85$ fixed. (b) Varying graph fusion weight $\gamma$ with $\lambda_{\mathrm{dis}}=0.10$ fixed.}
\label{fig:hyperparam}
\end{figure}

Fig.~\ref{fig:hyperparam} shows the sensitivity of the two key hyperparameters. For $\lambda_{\mathrm{dis}}$ (Fig.~\ref{fig:hyperparam}(a), $\gamma\!=\!0.85$ fixed), performance peaks at 0.1; smaller values under-regularize the decoupling, while larger values suppress useful private features (mIoU drops by 1.30\% at 1.0). For $\gamma$ (Fig.~\ref{fig:hyperparam}(b), $\lambda_{\mathrm{dis}}\!=\!0.1$ fixed), 0.85 balances spatial precision from base logits with relational context from graph reasoning. We therefore adopt $\lambda_{\mathrm{dis}}\!=\!0.1$ and $\gamma\!=\!0.85$ in all experiments.

\section{Conclusion}

This paper introduces URTF and GSCNet for unaligned UAV RGBT image semantic segmentation. URTF contains over 25,000 RGB-Thermal image pairs with 61 fine-grained semantic categories and realistic cross-modal misalignment, without pixel-level registration. GSCNet combines FDAM, which aligns modality-shared structural features through illumination-aware bidirectional deformable warping, with SGCM, which calibrates category predictions through taxonomy and co-occurrence priors via graph-attention reasoning. On URTF, GSCNet reaches 71.04\% mIoU and 60.17\% Tail-16 IoU.

GSCNet uses 160.66M parameters and runs at 16.62 FPS, which limits real-time UAV deployment. Future work will explore lighter variants through knowledge distillation and efficient backbones, as well as cross-resolution RGB-Thermal segmentation where the two modalities have different native resolutions.

\bibliographystyle{IEEEtran}

\bibliography{ref}

\begin{thebibliography}{10}
\providecommand{\url}[1]{#1}
\csname url@samestyle\endcsname
\providecommand{\newblock}{\relax}
\providecommand{\bibinfo}[2]{#2}
\providecommand{\BIBentrySTDinterwordspacing}{\spaceskip=0pt\relax}
\providecommand{\BIBentryALTinterwordstretchfactor}{4}
\providecommand{\BIBentryALTinterwordspacing}{\spaceskip=\fontdimen2\font plus
\BIBentryALTinterwordstretchfactor\fontdimen3\font minus \fontdimen4\font\relax}
\providecommand{\BIBforeignlanguage}[2]{{%
\expandafter\ifx\csname l@#1\endcsname\relax
\typeout{** WARNING: IEEEtran.bst: No hyphenation pattern has been}%
\typeout{** loaded for the language `#1'. Using the pattern for}%
\typeout{** the default language instead.}%
\else
\language=\csname l@#1\endcsname
\fi
#2}}
\providecommand{\BIBdecl}{\relax}
\BIBdecl

\bibitem{wang2021loveda}
J.~Wang, Z.~Zheng, A.~Ma, X.~Lu, and Y.~Zhong, ``Love{DA}: A remote sensing land-cover dataset for domain adaptive semantic segmentation,'' in \emph{Proc. Adv. Neural Inf. Process. Syst. (NeurIPS)}, 2021.

\bibitem{lyu2020uavid}
Y.~Lyu, G.~Vosselman, G.-S. Xia, A.~Yilmaz, and M.~Y. Yang, ``{UAVid}: A semantic segmentation dataset for {UAV} imagery,'' \emph{ISPRS J. Photogramm. Remote Sens.}, vol. 165, pp. 108--119, 2020.

\bibitem{sakaridis2018semantic}
C.~Sakaridis, D.~Dai, and L.~Van~Gool, ``Semantic foggy scene understanding with synthetic data,'' \emph{Int. J. Comput. Vis.}, vol. 126, no.~9, pp. 973--992, 2018.

\bibitem{vertens2020heatnet}
J.~Vertens, J.~Z{\"u}rn, and W.~Burgard, ``{HeatNet}: Bridging the day-night domain gap in semantic segmentation with thermal images,'' in \emph{Proc. IEEE/RSJ Int. Conf. Intell. Robots Syst. (IROS)}, 2020, pp. 8461--8468.

\bibitem{ha2017mfnet}
Q.~Ha, K.~Watanabe, T.~Karasawa, Y.~Ushiku, and T.~Harada, ``{MFNet}: Towards real-time semantic segmentation for autonomous vehicles with multi-spectral scenes,'' in \emph{Proc. IEEE/RSJ Int. Conf. Intell. Robots Syst. (IROS)}, 2017, pp. 5108--5115.

\bibitem{shivakumar2020pst900}
S.~S. Shivakumar, N.~Rodrigues, A.~Zhou, I.~D. Miller, V.~Kumar, and C.~J. Taylor, ``{PST900}: {RGB}-thermal calibration, dataset and segmentation network,'' in \emph{Proc. IEEE Int. Conf. Robot. Autom. (ICRA)}, 2020, pp. 9441--9447.

\bibitem{zhang2023cmx}
J.~Zhang, H.~Liu, K.~Yang, X.~Hu, R.~Liu, and R.~Stiefelhagen, ``{CMX}: Cross-modal fusion for {RGB-X} semantic segmentation with transformers,'' \emph{IEEE Trans. Intell. Transp. Syst.}, vol.~24, no.~12, pp. 14\,679--14\,694, 2023.

\bibitem{chen2024weakly}
C.~Chen, J.~Qi, X.~Liu, K.~Bin, R.~Fu, X.~Hu, and P.~Zhong, ``Weakly misalignment-free adaptive feature alignment for {UAV}s-based multimodal object detection,'' in \emph{Proc. IEEE/CVF Conf. Comput. Vis. Pattern Recognit. (CVPR)}, 2024, pp. 26\,836--26\,845.

\bibitem{liu2023multi}
J.~Liu, Z.~Liu, G.~Wu, L.~Ma, R.~Liu, W.~Zhong, Z.~Luo, and X.~Fan, ``Multi-interactive feature learning and a full-time multi-modality benchmark for image fusion and segmentation,'' in \emph{Proc. IEEE/CVF Int. Conf. Comput. Vis. (ICCV)}, 2023, pp. 8115--8124.

\bibitem{zhang2024mrfs}
H.~Zhang, X.~Zuo, J.~Jiang, C.~Guo, and J.~Ma, ``{MRFS}: Mutually reinforcing image fusion and segmentation,'' in \emph{Proc. IEEE/CVF Conf. Comput. Vis. Pattern Recognit. (CVPR)}, 2024, pp. 26\,974--26\,983.

\bibitem{chen2024cart}
C.~Chen \emph{et~al.}, ``{CART}: Cross-modal alignment for {RGB}-thermal semantic segmentation in {UAV} scenarios,'' in \emph{Proc. IEEE Int. Conf. Multimedia Expo (ICME)}, 2024.

\bibitem{ji2024unleashing}
W.~Ji, J.~Li, W.~Li, Y.~Shen, H.~Jin \emph{et~al.}, ``Unleashing multispectral video's potential in semantic segmentation: A semi-supervised viewpoint and new {UAV}-view benchmark,'' \emph{Adv. Neural Inf. Process. Syst.}, vol.~37, pp. 65\,717--65\,737, 2024.

\bibitem{ouyang2025kust4k}
J.~Ouyang, Q.~Wang, Y.~Shang, P.~Jin, H.~Zhong, L.~Zhou, and T.~Shen, ``An {RGB-TIR} dataset from {UAV} platform for robust urban traffic scenes semantic segmentation,'' \emph{Sci. Data}, 2025.

\bibitem{hazirbas2016fusenet}
C.~Hazirbas, L.~Ma, C.~Domokos, and D.~Cremers, ``{FuseNet}: Incorporating depth into semantic segmentation via fusion-based {CNN} architecture,'' in \emph{Proc. Asian Conf. Comput. Vis. (ACCV)}, 2016, pp. 213--228.

\bibitem{sun2019rtfnet}
Y.~Sun, W.~Zuo, and M.~Liu, ``{RTFNet}: {RGB}-thermal fusion network for semantic segmentation of urban scenes,'' \emph{IEEE Robot. Autom. Lett.}, vol.~4, no.~3, pp. 2576--2583, 2019.

\bibitem{deng2021feanet}
F.~Deng, H.~Feng, M.~Liang, H.~Wang, Y.~Yang, Y.~Gao, J.~Chen, J.~Hu, X.~Guo, and T.~L. Lam, ``{FEANet}: Feature-enhanced attention network for {RGB}-thermal real-time semantic segmentation,'' in \emph{Proc. IEEE/RSJ Int. Conf. Intell. Robots Syst. (IROS)}, 2021, pp. 4467--4473.

\bibitem{zhou2022edge}
W.~Zhou, S.~Dong, C.~Xu, and Y.~Qian, ``Edge-aware guidance fusion network for {RGB}--thermal scene parsing,'' in \emph{Proc. AAAI Conf. Artif. Intell.}, vol.~36, no.~3, 2022, pp. 3571--3579.

\bibitem{zhou2022gmnet}
W.~Zhou, J.~Liu, J.~Lei, L.~Yu, and J.-N. Hwang, ``{GMNet}: Graded-feature multilabel-learning network for {RGB}-thermal urban scene semantic segmentation,'' \emph{IEEE Trans. Image Process.}, vol.~30, pp. 7790--7802, 2021.

\bibitem{Zhong_2025_ICCV}
H.~Zhong, F.~Tang, Z.~Chen, H.~J. Chang, and Y.~Gao, ``{AMDANet}: Attention-driven multi-perspective discrepancy alignment for {RGB}-infrared image fusion and segmentation,'' in \emph{Proc. IEEE/CVF Int. Conf. Comput. Vis. (ICCV)}, October 2025, pp. 10\,645--10\,655.

\bibitem{yin2025dformerv2}
B.-W. Yin, J.-L. Cao, M.-M. Cheng, and Q.~Hou, ``{DFormerV2}: Geometry self-attention for {RGBD} semantic segmentation,'' in \emph{Proc. IEEE/CVF Conf. Comput. Vis. Pattern Recognit. (CVPR)}, 2025, pp. 19\,345--19\,355.

\bibitem{wang2026mambaseg}
F.~Gu, Y.~Li, X.~Long, K.~Ji, C.~Chen, Q.~Gu, and Z.~Ni, ``{MambaSeg}: Harnessing {Mamba} for accurate and efficient image-event semantic segmentation,'' in \emph{Proc. AAAI Conf. Artif. Intell.}, 2026.

\bibitem{zhou2025drgbt}
H.~Zhou, Z.~Zhang, C.~Li, C.~Tian, Y.~Xie, Z.~Li, and X.-J. Wu, ``Deformation-resilient multigranularity learning for unaligned {RGB}--{T} semantic segmentation,'' \emph{IEEE Trans. Neural Netw. Learn. Syst.}, vol.~36, no.~10, pp. 18\,530--18\,544, 2025.

\bibitem{long2015fully}
J.~Long, E.~Shelhamer, and T.~Darrell, ``Fully convolutional networks for semantic segmentation,'' in \emph{Proc. IEEE/CVF Conf. Comput. Vis. Pattern Recognit. (CVPR)}, 2015, pp. 3431--3440.

\bibitem{chen2017deeplab}
L.-C. Chen, G.~Papandreou, I.~Kokkinos, K.~Murphy, and A.~L. Yuille, ``{DeepLab}: Semantic image segmentation with deep convolutional nets, atrous convolution, and fully connected {CRF}s,'' \emph{IEEE Trans. Pattern Anal. Mach. Intell.}, vol.~40, no.~4, pp. 834--848, 2017.

\bibitem{yu2015multi}
F.~Yu and V.~Koltun, ``Multi-scale context aggregation by dilated convolutions,'' \emph{arXiv preprint arXiv:1511.07122}, 2015.

\bibitem{chen2018deeplabv3plus}
L.-C. Chen, Y.~Zhu, G.~Papandreou, F.~Schroff, and H.~Adam, ``Encoder-decoder with atrous separable convolution for semantic image segmentation,'' in \emph{Proc. Eur. Conf. Comput. Vis. (ECCV)}, 2018, pp. 801--818.

\bibitem{chen2019glore}
Y.~Chen, M.~Rohrbach, Z.~Yan, S.~Yan, J.~Feng, and Y.~Kalantidis, ``Graph-based global reasoning networks,'' in \emph{Proc. IEEE/CVF Conf. Comput. Vis. Pattern Recognit. (CVPR)}, 2019.

\bibitem{zhang2020dgmn}
L.~Zhang, D.~Xu, A.~Arnab, and P.~H. Torr, ``Dynamic graph message passing networks,'' in \emph{Proc. IEEE/CVF Conf. Comput. Vis. Pattern Recognit. (CVPR)}, 2020.

\bibitem{gui2025sagrnet}
B.~Gui, L.~Sam, A.~Bhardwaj, D.~S. G{\'o}mez, F.~G. Pe{\~n}aloza, M.~F. Buchroithner, and D.~R. Green, ``{SAGRNet}: A novel object-based graph convolutional neural network for diverse vegetation cover classification in remotely-sensed imagery,'' \emph{ISPRS J. Photogramm. Remote Sens.}, vol. 227, pp. 99--124, 2025.

\bibitem{bertinetto2020hierarchy}
L.~Bertinetto, R.~Mueller, K.~Tertikas, S.~Samangooei, and N.~A. Lord, ``Making better mistakes: Leveraging class hierarchies with deep networks,'' in \emph{Proc. IEEE/CVF Conf. Comput. Vis. Pattern Recognit. (CVPR)}, 2020.

\bibitem{chen2019mlgcn}
Z.-M. Chen, X.-S. Wei, P.~Wang, and Y.~Guo, ``Multi-label image recognition with graph convolutional networks,'' in \emph{Proc. IEEE/CVF Conf. Comput. Vis. Pattern Recognit. (CVPR)}, 2019, pp. 5177--5186.

\bibitem{detone2016deep}
D.~DeTone, T.~Malisiewicz, and A.~Rabinovich, ``Deep image homography estimation,'' \emph{arXiv preprint arXiv:1606.03798}, 2016.

\bibitem{jaderberg2015spatial}
M.~Jaderberg, K.~Simonyan, A.~Zisserman \emph{et~al.}, ``Spatial transformer networks,'' \emph{Adv. Neural Inf. Process. Syst.}, vol.~28, 2015.

\bibitem{dosovitskiy2015flownet}
A.~Dosovitskiy, P.~Fischer, E.~Ilg, P.~Hausser, C.~Hazirbas, V.~Golkov, P.~Van Der~Smagt, D.~Cremers, and T.~Brox, ``Flownet: Learning optical flow with convolutional networks,'' in \emph{Proc. IEEE Int. Conf. Comput. Vis. (ICCV)}, 2015, pp. 2758--2766.

\bibitem{sun2018pwc}
D.~Sun, X.~Yang, M.-Y. Liu, and J.~Kautz, ``{PWC-Net}: {CNNs} for optical flow using pyramid, warping, and cost volume,'' in \emph{Proc. IEEE/CVF Conf. Comput. Vis. Pattern Recognit. (CVPR)}, 2018, pp. 8934--8943.

\bibitem{zhu2019deformable}
X.~Zhu, H.~Hu, S.~Lin, and J.~Dai, ``Deformable convnets v2: More deformable, better results,'' in \emph{Proc. IEEE/CVF Conf. Comput. Vis. Pattern Recognit. (CVPR)}, 2019, pp. 9308--9316.

\bibitem{tip_e2e_reg_seg}
W.~Lai, F.~Zeng, X.~Hu, S.~He, Z.~Liu, and Y.~Jiang, ``{RegSeg}: An end-to-end network for multimodal {RGB}-thermal registration and semantic segmentation,'' \emph{IEEE Trans. Image Process.}, vol.~33, pp. 6676--6690, 2024.

\bibitem{hazarika2020misa}
D.~Hazarika, R.~Zimmermann, and S.~Poria, ``{MISA}: Modality-invariant and-specific representations for multimodal sentiment analysis,'' in \emph{Proc. 28th ACM Int. Conf. Multimedia}, 2020, pp. 1122--1131.

\bibitem{xu2020learning}
X.~Xu, K.~Lin, L.~Gao, H.~Lu, H.~T. Shen, and X.~Li, ``Learning cross-modal common representations by private--shared subspaces separation,'' \emph{IEEE Trans. Cybern.}, vol.~52, no.~5, pp. 3261--3275, 2020.

\bibitem{xie2021segformer}
E.~Xie, W.~Wang, Z.~Yu, A.~Anandkumar, J.~M. Alvarez, and P.~Luo, ``{SegFormer}: Simple and efficient design for semantic segmentation with transformers,'' \emph{Adv. Neural Inf. Process. Syst.}, vol.~34, pp. 12\,077--12\,090, 2021.

\bibitem{wang2019rgb}
G.~Wang, T.~Zhang, J.~Cheng, S.~Liu, Y.~Yang, and Z.~Hou, ``Rgb-infrared cross-modality person re-identification via joint pixel and feature alignment,'' in \emph{Proc. IEEE/CVF Int. Conf. Comput. Vis. (ICCV)}, 2019, pp. 3623--3632.

\bibitem{chen2020disentangled}
H.~Chen, Y.~Deng, Y.~Li, T.-Y. Hung, and G.~Lin, ``{RGBD} salient object detection via disentangled cross-modal fusion,'' \emph{IEEE Trans. Image Process.}, vol.~29, pp. 8407--8416, 2020.

\bibitem{velickovic2018graph}
P.~Veli{\v{c}}kovi{\'c}, G.~Cucurull, A.~Casanova, A.~Romero, P.~Li{\`o}, and Y.~Bengio, ``Graph attention networks,'' in \emph{Proc. Int. Conf. Learn. Represent. (ICLR)}, 2018.

\bibitem{rahnemoonfar2021floodnet}
M.~Rahnemoonfar, T.~Chowdhury, A.~Sarkar, D.~Varshney, M.~Yari, and R.~R. Murphy, ``{FloodNet}: A high resolution aerial imagery dataset for post flood scene understanding,'' \emph{IEEE Access}, vol.~9, pp. 89\,644--89\,654, 2021.

\bibitem{cai2025vdd}
W.~Cai, K.~Jin, J.~Hou, C.~Guo, L.~Wu, and W.~Yang, ``{VDD}: Varied drone dataset for semantic segmentation,'' \emph{J. Vis. Commun. Image Represent.}, 2025.

\bibitem{ji2023semanticrt}
W.~Ji, J.~Li, C.~Bian, Z.~Zhang, and L.~Cheng, ``{SemanticRT}: A large-scale dataset and method for robust semantic segmentation in multispectral images,'' in \emph{Proc. 31st ACM Int. Conf. Multimedia}, 2023, pp. 3307--3316.

\bibitem{sun2019hrnet}
K.~Sun, B.~Xiao, D.~Liu, and J.~Wang, ``Deep high-resolution representation learning for human pose estimation,'' in \emph{Proc. IEEE/CVF Conf. Comput. Vis. Pattern Recognit. (CVPR)}, 2019, pp. 5693--5703.

\bibitem{cheng2021mask2former}
B.~Cheng, I.~Misra, A.~G. Schwing, A.~Kirillov, and R.~Girdhar, ``Masked-attention mask transformer for universal image segmentation,'' in \emph{Proc. IEEE/CVF Conf. Comput. Vis. Pattern Recognit. (CVPR)}, 2022.

\bibitem{guo2022segnext}
M.-H. Guo, C.-Z. Lu, Q.~Hou, Z.~Liu, M.-M. Cheng, and S.-M. Hu, ``Seg{N}e{X}t: Rethinking convolutional attention design for semantic segmentation,'' in \emph{Proc. Adv. Neural Inf. Process. Syst. (NeurIPS)}, 2022.

\bibitem{Chen2024EdgeFE}
Z.~Chen, T.~Xu, Y.~Pan, N.~Shen, H.~Chen, and J.~Li, ``Edge feature enhancement for fine-grained segmentation of remote sensing images,'' \emph{IEEE Trans. Geosci. Remote Sens.}, vol.~62, pp. 1--13, 2024.

\bibitem{zhang2023delivering}
J.~Zhang, R.~Liu, H.~Shi, K.~Yang, S.~Rei{\ss}, K.~Peng, H.~Fu, K.~Wang, and R.~Stiefelhagen, ``Delivering arbitrary-modal semantic segmentation,'' in \emph{Proc. IEEE/CVF Conf. Comput. Vis. Pattern Recognit. (CVPR)}, 2023.

\bibitem{Wang2023sgfnet}
Y.~Wang, G.~Li, and Z.~Liu, ``{SGFNet}: Semantic-guided fusion network for {RGB}-thermal semantic segmentation,'' \emph{IEEE Trans. Circuits Syst. Video Technol.}, vol.~33, no.~12, pp. 7737--7748, 2023.

\bibitem{Shin2024ComplementaryRM}
U.~Shin, K.~Lee, and I.-S. Kweon, ``Complementary random masking for {RGB}-thermal semantic segmentation,'' in \emph{Proc. IEEE Int. Conf. Robot. Autom. (ICRA)}, 2024, pp. 11\,110--11\,117.

\bibitem{jia2024geminifusion}
D.~Jia, J.~Guo, K.~Han, H.~Wu, C.~Zhang, C.~Xu, and X.~Chen, ``Gemini{F}usion: Efficient pixel-wise multimodal fusion for vision transformer,'' in \emph{Proc. Int. Conf. Mach. Learn. (ICML)}, 2024.

\bibitem{ZHANG2024574asa}
P.~Zhang, B.~Peng, C.~Lu, Q.~Huang, and D.~Liu, ``{ASANet}: Asymmetric semantic aligning network for {RGB} and {SAR} image land cover classification,'' \emph{ISPRS J. Photogramm. Remote Sens.}, vol. 218, pp. 574--587, 2024.

\bibitem{milnet2025tip}
J.~Liu, H.~Liu, X.~Li, J.~Ren, and X.~Xu, ``{MiLNet}: Multiplex interactive learning network for {RGB-T} semantic segmentation,'' \emph{IEEE Trans. Image Process.}, vol.~34, pp. 1686--1699, 2025.

\bibitem{mulvmamba2026kbs}
R.~Ni, Y.~Guo, B.~Yang, Y.~Liu, H.~Wang, and C.~Hu, ``{Mul-VMamba}: Multimodal semantic segmentation using selection-fusion-based vision-{Mamba},'' \emph{Knowl.-Based Syst.}, vol. 334, p. 115119, 2026.

\end{thebibliography}

\end{document}